\documentclass{article}
\usepackage[margin=1.3in]{geometry}
\usepackage{times}
\usepackage{graphicx} 
\usepackage{subfigure} 

\usepackage{natbib}
\usepackage{amssymb, amsmath,amsthm,amsfonts}
\usepackage{amsfonts}
\usepackage{algorithm}
\usepackage{algorithmic}
\usepackage{paralist}
\usepackage{multirow}
\usepackage{times}
\usepackage{wrapfig}

\usepackage{url,enumerate}
\usepackage{color,xcolor}
\usepackage{epsfig,subfigure,wrapfig}
\usepackage{makeidx}  
\usepackage{amsmath,amssymb}
\usepackage[small, compact]{titlesec}
\usepackage{xspace}
\usepackage{epstopdf}
\usepackage{cite}
\usepackage{mathrsfs}
\usepackage{times}
\usepackage{enumerate}
\usepackage{color}
\usepackage{graphicx,epsfig,subfigure}
\usepackage{amsmath,amssymb,xspace}
\usepackage[breaklinks=true]{hyperref}
\usepackage{bm}
\usepackage{bbm}
\usepackage{upgreek}
\usepackage{cleveref}
\usepackage{chngcntr}

\usepackage{hyperref}

\newtheorem{theorem}{Theorem}[section]
\newtheorem{lemma}[theorem]{Lemma}

\newcommand{\ie}{{{i.e.,}}\xspace}
\newcommand{\eg}{{{\em e.g.,}}\xspace}
\newcommand{\cmt}[1]{}
\newcommand{\ours}{{{ADVGP}}\xspace}

\newcommand{\mapreduce}{\textsc{MapReduce}\xspace}

\newcommand{\mappers}{\textsc{Mappers}\xspace}

\newcommand{\reducers}{\textsc{Reducers}\xspace}

\newcommand{\alanc}[1]{}

\newcommand{\ps}{\textsc{ParameterServer}\xspace}

\newcommand{\diag}{{\rm diag}}

\renewcommand{\d}{{\rm d}}  

\newcommand{\f}{{\bf f}}

\renewcommand{\k}{{\bf k}}

\newcommand{\w}{{\bf w}}
\newcommand{\x}{{\bf x}}
\newcommand{\y}{{\bf y}}
\newcommand{\z}{{\bf z}}

\newcommand{\I}{{\bf I}}

\newcommand{\K}{{\bf K}}
\renewcommand{\L}{{\bf L}}

\newcommand{\N}{\mathcal{N}}  

\newcommand{\Q}{{\bf Q}}

\newcommand{\U}{{\bf U}}

\newcommand{\X}{{\bf X}}

\newcommand{\Z}{{\bf Z}}

\newcommand{\balpha}{\boldsymbol{\alpha}}

\newcommand{\bphi}{\boldsymbol{\phi}}

\newcommand{\Beta}{\boldsymbol{\eta}}

\newcommand{\blambda}{\boldsymbol{\lambda}}

\newcommand{\btheta}{\boldsymbol{\theta}}

\newcommand{\bSigma}{\boldsymbol{\Sigma}}

\newcommand{\bPhi}{\boldsymbol{\Phi}}

\newcommand{\bmu}{\boldsymbol{\mu}}

\newcommand{\0}{{\bf 0}}

\newcommand{\ben}{\begin{enumerate}}
\newcommand{\een}{\end{enumerate}}

\title{Asynchronous Distributed Variational  Gaussian Process for Regression}

\author{
	Peng, Hao\\
	\texttt{pengh@purdue.edu}
	\and
	Zhe, Shandian\\
	\texttt{szhe@purdue.edu}
	\and
	Qi, Yuan \\
	\texttt{alan0@outlook.com}
}

\begin{document} 
\maketitle

\begin{abstract} 
 Gaussian processes (GPs) are powerful non-parametric function estimators. However, their applications are largely limited by the expensive computational cost  of the inference procedures. Existing stochastic or distributed synchronous variational inferences, although have alleviated this issue by scaling up GPs to millions of samples, are still far from satisfactory for real-world large applications, where the data sizes are often orders of magnitudes larger, say, billions. To solve this problem, we propose \ours, the first \underline{A}synchronous \underline{D}istributed \underline{V}ariational \underline{G}aussian \underline{P}rocess inference for regression,  on the recent large-scale machine learning platform, \ps. \ours uses a novel, flexible variational framework based on a weight space augmentation, and implements the highly efficient, asynchronous proximal gradient optimization. While maintaining comparable or better predictive performance, \ours greatly improves upon the efficiency of the existing variational methods. With \ours, we effortlessly scale up GP regression to a real-world application with billions of samples and demonstrate an excellent, superior prediction accuracy to the popular linear models. 
 
\end{abstract} 
\section{Introduction}

Gaussian processes (GPs)~\citep{GPML06} are powerful non-parametric Bayesian models for function estimation. Without imposing any explicit parametric form, GPs merely induce a smoothness assumption via the definition of covariance function, and hence can flexibly infer various, complicated functions from data. In addition, GPs are robust to noise, resist overfitting and produce uncertainty estimations.  
However, a crucial bottleneck of GP models is their expensive computational cost: exact GP inference requires $\mathcal{O}(n^3)$ time complexity and $\mathcal{O}(n^2)$ space complexity ($n$ is the number of training samples), which limits GPs to very small applications, say, a few hundreds of samples. 

To mitigate this limitation, many approximate inference algorithms have been developed~\citep{williams2001using, seeger2003fast, quinonero2005unifying,snelson2005sparse,deisenroth2015distributed}. 
Most methods use sparse approximations. Basically, we first introduce a small set of inducing points; and then  we develop an approximation that transfers the expensive computations from the entire large training data, such as the covariance and inverse covariance matrix calculations, to the small set of the inducing points.  To this end, a typical strategy is to impose some simplified modeling assumption. For example, FITC~\citep{snelson2005sparse} makes a fully conditionally independent assumption. Recently, \citet{titsias2009variational} proposed a more principled, variational sparse approximation framework, where the inducing points are also treated as variational parameters. The variational framework is less prone to overfitting and often yields a better inference quality~\citep{titsias2009variational,bauer2016understanding}. Based on the variational approximation, \citet{GPSVI13} developed a stochastic variational inference (SVI) algorithm, and \citet{gal2014distributed} used a tight variational lower bound to develop a distributed inference algorithm with the \mapreduce framework. 

While SVI and the distributed variational inference have successfully scaled up GP models to millions of samples ($\mathcal{O}(10^6)$), they are still insufficient for real-world large-scale applications, in which the data sizes are often orders of magnitude larger, say, over billions of samples~($\mathcal{O}(10^9)$). Specifically, SVI~\citep{GPSVI13} sequentially processes data samples and requires too much time to complete even one epoch of training. The distributed variational algorithm in~\citep{gal2014distributed} uses the \mapreduce framework and requires massive synchronizations during training, where a large amount of time is squandered when the \mappers or \reducers are waiting for each other, or the failed nodes are restarted.  

To tackle this problem, we propose \underline{A}synchronous \underline{D}istributed \underline{V}ariational \underline{G}aussian \underline{P}rocess inference (\ours), which enables GP regression on applications with (at least) billions of samples. To the best of our knowledge, this is the first variational inference that scales up GPs to this level. 
The contributions of our work are summarized as follows: first, we propose a novel, general variational GP framework using a weight space augmentation (Section 3). The framework allows flexible constructions of feature mappings to incorporate various low-rank structures and to fulfill different variational model evidence lower bounds (ELBOs). Furthermore, due to the simple standard normal prior of the random weights, the framework enables highly efficient, asynchronous proximal gradient-based optimization, with convergence guarantees as well as fast, element-wise and parallelizable variational posterior updates.
Second, based on the new framework, we develop a highly efficient, asynchronous variational inference algorithm in the recent distributed machine learning platform, \ps~\citep{li2014scaling} (Section 4). The asynchronous algorithm eliminates an enormous amount of waiting time caused by the synchronous coordination, and fully exploits the computational power and network bandwidth; as a result, our new inference, \ours, greatly improves on both the scalability and  efficiency of the prior variational algorithms, while still maintaining a similar or better inference quality. Finally, in a real-world application with billions of samples, we effortlessly train a GP regression model with \ours and achieve an excellent prediction accuracy, with $17\%$ improvement over the popular linear regression implemented in Vowpal Wabbit~\citep{agarwal14}, the  state-of-the-art large-scale machine learning software widely used in industry. 
\section{Gaussian Processes Review}\label{sect:GP-review}
In this paper, we focus on Gaussian process (GP) regression. Suppose we aim to infer an underlying function $f:\mathbb{R}^d \rightarrow \mathbb{R}$ from an observed dataset $\mathcal{D}=\{\X, \y\}$, where $\X=[\x_1^\top, \ldots, \x_n^\top]^\top$ is the input matrix and $\y$ is the output vector. Each row of $\X$, namely $\x_i$ ($1 \le i \le n$),  is a $d$-dimensional input vector. Correspondingly, each element of $\y$, namely $y_i$,  is an observed function value corrupted by some random noise. Note that the function $f$ can be highly nonlinear. To estimate $f$ from $\mathcal{D}$, we place a GP prior over $f$. Specifically, we treat the collection of all the function values as one realization of the Gaussian process. Therefore, the finite projection of $f$ over the inputs $\X$, \ie $\f = [f(\x_1), \ldots, f(\x_n)]$ follows a multivariate Gaussian distribution: 
$\f \sim \N\big(\f|\bar{\f}, \K_{nn}\big)$, where $\bar{\f}=[\bar{f}(\x_1), \ldots, \bar{f}(\x_n)]$ are the mean function values and $\K_{nn}$ is the $n \times n$ covariance matrix. 
Each element of  $\K_{nn}$ is a covariance function $k(\cdot, \cdot)$ of two input vectors, \ie $[\K_{nn}]_{i,j} = k(\x_i, \x_j)$. We can choose any symmetric positive semidefinite kernel as the covariance function, \eg the ARD kernel:
$k(\x_i, \x_j) = a_0^2 \exp\left(-\frac{1}{2} (\x_i-\x_j)^\top \diag(\Beta) (\x_i -\x_j)\right)$,
where $\Beta = [1/a_1^2, ..., 1/a_d^2]$.
For simplicity, we usually use the zero mean function, namely $\bar{f}(\cdot)=0$.  

Given $\f$, we use an isotropic Gaussian model to sample the observed noisy output $\y$: $p(\y|\f) = \N(\y|\f, \beta^{-1}\I)$. The joint probability of  GP regression is  
\begin{align}
p(\y,\f|\X) = \N\big(\f|\0, \K_{nn}\big) \N(\y|\f, \beta^{-1}\I). \label{GPR}
\end{align}
Further, we can obtain the marginal distribution of $\y$, namely the model evidence, by marginalizing out $\f$:
\begin{align}
p(\y|\X) = \N\big(\y|\0, \K_{nn} + \beta^{-1}\I\big).\label{GPR-M}
\end{align}
The inference of GP regression aims to estimate the appropriate kernel parameters and noise variance from the training data $\mathcal{D} = \{\X, \y\}$, such as $\{a_0, \Beta\}$ in ARD kernel  and $\beta^{-1}$. To this end, we can maximize the model evidence in \eqref{GPR-M} with respect to those parameters.  However,  to maximize  \eqref{GPR-M}, we need to calculate the inverse and the determinant of the $n \times n$ matrix $\K_{nn} + \beta^{-1}\I$ to evaluate the multivariate Gaussian term. This will take  $\mathcal{O}(n^3)$ time complexity and  $\mathcal{O}(n^2)$ space complexity and hence is infeasible for a large number of samples, \ie large $n$.

For prediction, given a test input $\x^*$, since the test output $f^*$ and training output $\f$ can be treated as another GP projection on $\X$ and $\x^*$, the joint distribution of $f^*$ and $\f$ is also a multivariate Gaussian distribution. 
Then by marginalizing out $\f$, we can obtain the posterior distribution of $f^*$: 
\begin{align}
p(f^*|\x^*, \X,\y) = \N(f^* | \alpha, v) \label{GPR-pred},
\end{align}
where
\begin{align}
\alpha &= \k_{n*}^\top (\K_{nn} + \beta^{-1} \I)^{-1}\y, \label{pred-mean}\\
v &= k(x^*, x^*) - \k_{n*}^\top(\K_{nn} + \beta^{-1}\I)^{-1}\k_{n*}, \label{pred-var}
\end{align}
and $\k_{n*} = [k(\x^*, \x_1), \ldots, k(\x^*, \x_n)]^\top$.
Note that the calculation also requires the inverse of $K_{nn} + \beta^{-1} \I$ and hence takes $\mathcal{O}(n^3)$ time complexity and  $\mathcal{O}(n^2)$ space complexity.

\section{Variational Framework Using Weight Space Augmentation} 
Although GPs  allow flexible function inference, 
they have a severe computational bottleneck. The training and prediction both require  $\mathcal{O}(n^3)$ time complexity and $\mathcal{O}(n^2)$ space complexity (see  \eqref{GPR-M},  \eqref{pred-mean} and  \eqref{pred-var}), making GPs unrealistic for real-world, large-scale applications, where the number of samples (\ie $n$) are often  billions or even larger. To address this problem,  we propose \ours that performs highly efficient, asynchronous distributed variational inference and enables the training of GP regression on extremely large data. \ours is based on a novel variational GP framework using a weight space augmentation, which is discussed below. 
 
First, we construct an equivalent augmented model by introducing an $m \times 1$ auxiliary random weight vector $\w$ ($m \ll n$). We assume $\w$ is sampled from the standard normal prior distribution: $p(\w) = \N(\w|\0, \I)$. Given $\w$, we sample an $n \times 1$ latent function values $\f$ from
\begin{align}
p(\f|\w)=\N(\f|\bPhi\w, \K_{nn} - \bPhi\bPhi^\top) \label{tf-sample},
\end{align}
where $\bPhi$ is an $n \times m$ matrix: $\bPhi=[\bphi(\x_1), \ldots, \bphi(\x_n)]^\top$. Here $\bphi(\cdot)$ represents a feature mapping that maps the original $d$-dimensional input into an $m$-dimensional feature space. Note that we need to choose an appropriate $\bphi(\cdot)$ to ensure the covariance matrix in  \eqref{tf-sample} is symmetric positive semidefinite. Flexible constructions of $\bphi(\cdot)$ enable us to fulfill different variational model evidence lower bounds (ELBO) for large-scale inference, which we will discuss more in Section \ref{sec:discussion}.

Given $\f$, we sample the observed output $\y$ from the isotropic Gaussian model $p(\y|\f)=\N(\y|{\f}, \beta^{-1}\I)$. The joint distribution of our augmented model is then given by
\begin{align}
&p(\y, {\f}, \w|\X) \nonumber \\
=&\N(\w|\0, \I)\N(\f|\bPhi\w, \K_{nn} - \bPhi\bPhi^\top)\N(\y|{\f}, \beta^{-1}\I). \label{GPR-W}
\end{align}
This model is equivalent to the original GP regression---when we marginalize out $\w$,  we recover the  joint distribution in  \eqref{GPR}; we can further marginalize out $\f$ to recover the model evidence in  \eqref{GPR-M}. Note that our model is distinct from  the traditional weight space view of GP regression~\citep{GPML06}: the feature mapping $\bphi(\cdot)$ is not equivalent to the underlying (nonlinear) feature mapping induced by the covariance function (see more discussions in Section \ref{sec:discussion}). Instead, $\bphi(\cdot)$ is defined for computational purpose only---that is, to construct a tractable variational evidence lower bound (ELBO),  shown as follows. 

Now,  we derive the tractable ELBO based on the weight space augmented model in  \eqref{GPR-W}. The derivation is similar to~\citep{titsias2009variational, GPSVI13}. 
Specifically, we first consider the conditional distribution $p(\y|\w)$. Because 
$\log p(\y|\w) = \log \int p(\y|\f) p(\f|\w)\d \f = \log \langle p(\y|\f) \rangle_{p(\f|\w)} $, 
where $\langle \cdot \rangle_{p(\btheta)}$ denotes the expectation under the distribution $p(\btheta)$, we can use Jensen's inequality to obtain a lower bound: 
\begin{align}
&\log p(\y|\w) = \log \langle p(\y|\f) \rangle_{p(\f|\w)} \ge  \langle \log p( \y | \f) \rangle_{p(\f|\w )} \nonumber \\
&= \sum_{i=1}^{n} \log \mathcal{N}(y_i | \bphi^\top(\x_i) \w, \beta^{-1})
- \frac{\beta}{2} \tilde{k}_{ii} \label{bound-middle},
\end{align}
where $\tilde{k}_{ii}$ is the $i$th diagonal element of $\K_{nn} - \bPhi\bPhi^\top$. 

Next, we introduce a variational posterior $q(\w)$ to construct the variational lower bound of the log model evidence,
\begin{align}
\log p(\y) &= \log \left\langle \frac{p(\y|\w)p(\w)}{q(\w)} \right\rangle_{q(\w)} \nonumber \\
&\ge   \langle \log p(\y|\w) \rangle_{q(\w)} - \mathrm{KL}(q(\w) \| p(\w)). \label{bound-1}
\end{align}
where $\mathrm{KL(\cdot \| \cdot)}$ is the Kullback--Leibler divergence. Replacing $\log p(\y|\w)$ in \eqref{bound-1} by the right side of \eqref{bound-middle}, we obtain the following lower bound,
\begin{align}
&\log p(\y) \ge \mathcal{L} = -\mathrm{KL}\left(q(\w) \| p(\w)\right) \nonumber \\
& + \sum_{i=1}^{n} \left \langle \log \mathcal{N}(y_i | \bphi^\top(\x_i)\w , \beta^{-1}) \right \rangle_{q(\w)}
- \frac{\beta}{2} \tilde{k}_{ii}. \label{bound-full}
\end{align}
Note that this is a variational lower bound: the equality is obtained when $\bPhi \bPhi^T=\K_{nn}$ and $q(\w) = p(\w|\y)$. To achieve equality, we need to set $m=n$ and have  $\bphi(\cdot)$ map the $d$-dimensional input into an $n$-dimensional feature space.  In order to reduce the computational cost, however, we can restrict $m$ to be very small and choose any family of mappings $\bphi(\cdot)$ that satisfy $\K_{nn} - \bPhi\bPhi^\top \succeq \0$. The flexible choices of $\bphi(\cdot)$ allows us to explore different approximations in a unified variational framework. For example, in our practice, 
we  introduce an $m \times d$ inducing matrix  $\Z = [\z_1, \ldots, \z_m ]^\top$ and  define
\begin{align}
\bphi(\x) =  \L^\top \k_m(\x) \label{mapping-1},
\end{align}
where $\k_m(\x) = [k(\x, \z_1), \ldots, k(\x,\z_m)]^\top$  and $\L$ is the lower triangular Cholesky factorization of the inverse kernel matrix over $\Z$, \ie  $[\K_{mm}]_{i,j} = k(\z_i, \z_j)$ and $\K_{mm}^{-1} = \L\L^\top$. It can be easily verified that $\bPhi \bPhi^\top = \K_{nm}\K_{mm}^{-1}\K_{nm}^\top$, where $\K_{nm} $ is the cross kernel matrix between $\X$ and $\Z$, \ie  $[\K_{nm}]_{ij}  = k(\x_i, \z_j)$. Therefore $\K_{nn} - \bPhi\bPhi$ is always positive semidefinite, because it can be viewed as a Schur complement of $\K_{nn}$ in the block matrix $\left[ \begin{smallmatrix} \K_{mm} \quad\K_{nm}^{\top}\\ \K_{nm} \quad\K_{nn} \end{smallmatrix}\right]$.
We discuss other choices of $\bphi(\cdot)$ in Section 5.                                                         

\section{Delayed Proximal Gradient Optimization for ADVGP}
A major advantage of our variational GP framework is the  capacity of using the  asynchronous, delayed proximal gradient optimization supported by \ps~\citep{li2014communication}, with convergence guarantees and scalability to huge data. \ps is a well-known, general platform for asynchronous machine learning algorithms for extremely large applications. It has a bipartite architecture where the computing nodes are partitioned into two classes: server nodes store the model parameters and worker nodes the data. \ps assumes the learning procedure minimizes a non-convex loss function with the following composite form:
\begin{align}
L(\btheta) = \sum\nolimits_{k=1}^{r} G_k(\btheta) + h(\btheta) \label{ps-obj}
\end{align}
where $\btheta$ are the model parameters. Here $G_k(\btheta)$ is a (possibly non-convex) function associated with the data in worker $k$ and therefore can be calculated by worker $k$ independently; $h(\btheta)$ is a convex function with respect to $\btheta$.

To efficiently minimize the loss function in  \eqref{ps-obj}, \ps uses a delayed proximal gradient updating method to perform asynchronous optimization. To illustrate it, let us first review the standard proximal gradient descent. Specifically,  for each iteration $t$, we first take a gradient descent step according to  $\sum\nolimits_k G_k(\btheta)$ and then perform a proximal operation to project $\btheta$ toward the minimum of $h(\cdot)$, \ie $\boldsymbol{\theta}^{(t+1)} = \mathrm{Prox}_{\gamma_t}[\btheta^{(t)} - \gamma_t \sum\nolimits_k \nabla G_k(\btheta^{(t)}) ]$, where $\gamma_t$ is the step size. The proximal operation is defined as 
\begin{align}
\mathrm{Prox}_{\gamma_t}[\btheta] =  \underset{{\btheta}^{*}}{\mathrm{argmin}} \;\;\; h(\btheta^{*}) + \frac{1}{2\gamma_t} \| \btheta^{*} - {\btheta} \|_2^2 \label{prox-op}. 
\end{align}
The standard proximal gradient descent guarantees to find a local minimum solution. However, the computation is inefficient, even in parallel: in each iteration, the server nodes wait until the worker nodes finish calculating  each $\nabla G_k(\btheta^{(t)})$; then the workers wait for the servers to finish the proximal operation. This synchronization wastes much time and computational resources. To address this issue, \ps uses a delayed proximal gradient updating approach to implement asynchronous computation. 


Specifically, we set a delay limit $\tau \ge 0$. At any iteration $t$, the servers do not enforce all the workers to finish iteration $t$; instead, as long as each worker has finished an iteration no earlier than  $t-\tau$, the servers will proceed to perform the proximal updates, \ie $\boldsymbol{\theta}^{(t+1)} = \mathrm{Prox}_{\gamma_t}[\btheta^{(t)} - \gamma_t \sum\nolimits_k \nabla G_k(\btheta^{(t_k)}) ]$  ($t-\tau \le t_k \le t$),  and notify all the workers with the new parameters $\boldsymbol{\theta}^{(t+1)}$. Once received the updated parameters, the workers compute and push the local gradient to the servers immediately.    
Obviously, this delay mechanism can effectively reduce the wait between the server and worker nodes.  By setting different  $\tau$, we can adjust the degree of the asynchronous computation: when $\tau = 0$, we have no asynchronization and return to the standard, synchronous proximal gradient descent; when $\tau = \infty$, we are totally asynchronous and there is no wait at all. 

A highlight is that given the composite form of the non-convex loss function in  \eqref{ps-obj}, the above asynchronous delayed proximal gradient descent guarantees to converge according to Theorem \ref{th1}.
\begin{theorem} \citep{li2013distributed} \label{th1}
	Assume the gradient of the function $G_k$ is Lipschitz continuous,
	that is, there is a constant $C_k$ such that $\|\nabla G_k(\boldsymbol{\theta})  -  \nabla G_k(\boldsymbol{\theta}') \| \leq C_k || \boldsymbol{\theta} - \boldsymbol{\theta}'||$ for any $\boldsymbol{\theta}$, $\boldsymbol{\theta}'$, and $k = 1,...,r$.
	Define $C = \sum_{k=1}^{r} C_k$. 
	Also, assume we allow a maximum delay for the updates by $\tau$ and a significantly-modified filter on pulling the parameters with threshold $\mathcal{O}(t^{-1})$. 
	For any $\epsilon > 0$, the delayed proximal gradient descent converges to a stationary point if the learning rate $\gamma_t$ satisfies $\gamma_t \le ((1+\tau)C +\epsilon)^{-1}$. \label{thm:convergence}
\end{theorem} 


Now, let us return to our variational GP framework. A major benefit of our framework is that the negative variational evidence lower bound  (ELBO) for GP regression has the same composite form as  \eqref{ps-obj}. Thereby we can  apply the asynchronous proximal gradient descent for GP inference on \ps. Specifically, we explicitly assume $q(\w) = \N(\w|\bmu, \bSigma)$ and obtain the negative  variational ELBO (see  \eqref{bound-full})
\begin{align}
\mathcal{-L} =  \sum\nolimits_{i=1}^n g_i + h 
\end{align}
where
\begin{align}
g_i & = -\log \N(y_i | \bphi^{\top}(\x_i)\bmu, \beta^{-1}) + \frac{\beta}{2} \bphi^{\top}(\x_i) \bSigma \bphi(\x_i) +\frac{\beta}{2}\tilde{k}_{ii}, \nonumber \\
h &  =\frac{1}{2} \left(-\ln |\boldsymbol{\Sigma}| - m + \mathrm{tr}(\bSigma) + \bmu^\top \bmu \right). 
\end{align}
Instead of directly updating $\bSigma$, we consider $\U$, the upper triangular Cholesky factor of $\bSigma$, \ie $\bSigma = \U^\top \U$. This not only simplifies the proximal operation but also ensures the positive definiteness of $\bSigma$ during computation. The partial derivatives of $g_i$ with respect to $\bmu$ and $\U$ are
\begin{align}
\frac{\partial g_i}{\partial \bmu}  &= \beta \left(-y_i \bphi(\x_i) +\bphi(\x_i)\bphi^\top(\x_i)\bmu \right), \\
\frac{\partial g_i}{\partial \U}  &= \beta\mathrm{triu}[\U \bphi(\x_i)\bphi^\top(\x_i)], 
\end{align}
where $\mathrm{triu}[\cdot]$ denotes the operator that keeps the upper triangular part of a matrix but leaves any other element zero.It can be verified that the partial derivatives of $g_i$ with respect to $\bmu$ and $\U$ are Lipschitz continuous and $h$ is also convex with respect to $\bmu$ and $\U$. According to Theorem \ref{th1}, minimizing $\mathcal{-L}$ (\ie maximizing $\mathcal{L}$) with respect to the variational parameters, $\bmu$ and $\U$, using the asynchronous proximal gradient method can guarantee convergence.  For other parameters, such as kernel parameters and inducing points, $h$ is simply a constant. As a result, the delayed proximal updates for these parameters reduce to the delayed gradient descent optimization such as in \citep{agarwal2011distributed}.


We now present the details of  \ours implementation on \ps. We first partition the data for $r$ workers and allocate the model parameters (such as the kernel parameters, the parameters of $q(\w)$ and the inducing points $\Z$) to server nodes.  At any iteration $t$, the server nodes  aggregate all the local gradients and  perform the proximal operation in  \eqref{prox-op},  as long as each worker $k$ has computed and pushed the local  gradient on its own data subset $D_k$ for some prior iteration $t_k$ ($t-\tau \le t_k \le t$),  $\nabla G_k^{(t_k)} =  \sum_{i \in D_k}\nabla g_i^{(t_k)}$. Note that the proximal operation is only performed for the parameters of $q(\w)$, namely $\bmu$ and $\U$;  since $h$ is constant for  the other model parameters, such as the kernel parameters and the inducing points, their gradient descent updates remain unchanged. Minimizing  \eqref{prox-op} by setting the derivatives to zero, we obtain the proximal updates for each element in $\bmu$ and $\U$:

\begin{align}
\mu^{(t+1)}_i &= \mu'^{(t+1)}_i / (1+\gamma_t) \label{eq:updMu}, \\
U^{(t+1)}_{ij} & = U'^{(t+1)}_{ij} / (1+\gamma_t) \label{eq:updSig}, \\
U^{(t+1)}_{ii} & = \frac{U'^{(t+1)}_{ii}+\sqrt{(U'^{(t+1)}_{ii})^2 + 4(1+\gamma_t)\gamma_t}}{2(1+\gamma_t)} \label{eq:updDiagSig},
\end{align}
where 
\begin{align} 
\mu'^{(t+1)}_i &= \mu^{(t)}_i - \gamma_t \sum\nolimits_{k=1}^r \frac{\partial G_k^{(t_k)}}{\partial \mu^{(t_k)}_i}, \nonumber \\
U'^{(t+1)}_{ij} & = U^{(t)}_{ij} - \gamma_t \sum\nolimits_{k=1}^r \frac{\partial G_k^{(t_k)}}{\partial U^{(t_k)}_{ij}}.   \nonumber
\end{align}
The proximal operation comprises of  element-wise, closed-form computations, therefore making the updates of the variational posterior $q(\w)$ highly parallelizable  and efficient. The gradient calculation for the other parameters, including the kernel parameters and inducing points, although quite complicated, is pretty standard and we give the details in the supplementary material (Appendix A). 
Finally, \ours is summarized in Algorithm \ref{alg:prox}. 
\begin{algorithm}
	\caption{Delayed Proximal Gradient for \ours}
	\label{alg:prox}
	\textbf{\underline{Worker $k$ at iteration $t_k$}}
	\begin{algorithmic}[1]
		\STATE Block until servers have new parameters ready.
		\STATE Pull the parameters from servers and update the current version (or iteration) $t_k$.
		\STATE Compute the gradient $\nabla G_k^{(t_k)}$ on data $D_k$. 
		\STATE Push the gradient $\nabla G_k^{(t_k)}$ to servers.
	\end{algorithmic}
	\textbf{\underline{Servers at iteration $t$}} 
	\begin{algorithmic}[1]
		\IF  {Each worker k completes iteration $t_k \ge t - \tau$}
		\STATE Aggregate gradients to obtain $\nabla G^{(t)} = \sum \nabla G_k^{(t_k)}$.
		\STATE Update $\boldsymbol{\mu}$ and $\boldsymbol{U}$ using (\ref{eq:updMu}), (\ref{eq:updSig}) and (\ref{eq:updDiagSig}).
		\STATE Update the other parameters using gradient descent.
		\STATE Notify all blocked workers of the new parameters and the version (\ie $t+1$).
		\STATE Proceed to iteration $t+1$.
		\ENDIF
	\end{algorithmic}
\end{algorithm}


\section{Discussion and Related Work} \label{sec:discussion}
Exact GP inference requires computing the full covariance matrix (and its inverse), and therefore is infeasible for large data. To reduce the computational cost, many sparse GP  inference methods use a low-rank structure to approximate the full covariance.  For example, 
\citet{williams2001using,hao2015eigengp} used the Nystr\"om approximation;  
\citet{BishopTipping00} used relevance vectors, constructed from covariance functions evaluated on a  small subset of the training data. 
A popular family of sparse GPs  introduced a small set of inducing inputs and targets,  viewed as statistical summary of the data, and define an approximate model by imposing some conditional independence between latent functions given the inducing targets; the inference of the inexact model is thereby much easier.  \citet{quinonero2005unifying} provided a unified view of those methods,
such as SoR~\citep{smola2001sparse}, DTC~\citep{seeger2003fast}, PITC~\citep{schwaighofer2003transductive} and FITC~\citep{snelson2005sparse}. 

Despite the success of those methods,  their inference procedures often exhibit  undesirable behaviors, such as underestimation of the noise and clumped inducing inputs \citep{bauer2016understanding}. To obtain a more favorable approximation, \citet{titsias2009variational} proposed a variational sparse GP framework, where the  approximate posteriors and the inducing inputs are both treated as variational parameters and estimated by maximizing a variational lower bound of the true model evidence. The variational framework is less prone to overfitting and often yields a better inference quality~\citep{titsias2009variational,bauer2016understanding}. Based on \citet{titsias2009variational}'s work, \citet{GPSVI13} developed a stochastic variational inference for GP (SVIGP) by parameterizing the variational distributions explicitly.
\citet{gal2014distributed} reparameterized the bound of \citet{titsias2009variational} and  developed a distributed optimization algorithm with  \mapreduce framework. Further,  \citet{dai2014gaussian} developed a GPU acceleration using the similar formulation, and  \citet{matthews2017gpflow} developed GPflow library, a TensorFlow implementation that exploit GPU hardwares. 

To further enable GPs on real-world, extremely large applications,  we proposed a new variational GP framework using a weight space augmentation.  The proposed augmented model, introducing an extra random weight vector $\w$ with standard normal prior,  is distinct from  the traditional GP weight space view~\citep{GPML06} and the recentering tricks used in GP MCMC inferences~\citep{murray2010slice, filippone2013comparative,  hensman2015mcmc}. In the conventional GP weight space view, the weight vector is used to combine the nonlinear feature mapping induced by the covariance function and therefore can be infinite dimensional; in the recentering tricks, the weight vector is used to reparameterize the latent function values, to dispose of the dependencies on the hyper-parameters, and to improve the mixing rate. 
In our framework, however, the weight vector $\w$ has a fixed, much smaller dimension than the number of samples ($m \ll n$), and is 
 used to introduce an extra feature mapping $\bphi(\cdot)$ --- $\bphi(\cdot)$ plays the key role to construct a tractable variational model evidence lower bound (ELBO) for large scale GP inference.  

The advantages of our framework are mainly twofold. First, by using the feature mapping $\bphi(\cdot)$, we are flexible to incorporate  various low rank structures, and meanwhile still cast them into a principled variational inference framework.  
For example, in addition to \eqref{mapping-1},  we can define 
\begin{align}
\bphi(\x) =  {\diag(\blambda)}^{-1/2}\Q^\top \k_{m}(\x), \label{eq:nystrom}
\end{align}
where $\Q$ are $\blambda$ are eigenvectors and eigenvalues of $\K_{mm}$. Then $\bphi(\cdot)$ is actually a scaled Nystr\"om approximation for eigenfunctions of the kernel used in GP regression. This  actually fulfills a variational version of the EigenGP approximation~\citep{hao2015eigengp}. Further, we can extend  \eqref{eq:nystrom} by combining  $q$ Nystr\"om approximations. Suppose we have $q$ groups of inducing inputs $\{\Z_1, \ldots, \Z_q\}$, where each $\Z_l$ consists of $m_l$ inducing inputs. Then the feature mapping can be defined by
\begin{align}
\bphi(\x) =  \sum^{q}_{l=1} q^{-1/2} {\diag(\blambda_l)}^{-1/2}\Q_l^\top \k_{m_l}(\x), \label{eq:ensemble_nystrom}
\end{align}
where $\blambda_l$ and $\Q_l$ are the eigenvalues and eigenvectors of the covariance matrix for $\Z_l$. This leads to a variational sparse GP based on the ensemble Nystr\"om approximation~\citep{kumar2009ensemble}. It can be trivially verified that both \eqref{eq:nystrom} and \eqref{eq:ensemble_nystrom} satisfied $\K_{nn} - \bPhi\bPhi^\top \succeq \0$ in \eqref{tf-sample}. 

In addition, we can also relate \ours to GP models with pre-defined feature mappings, for instance, 
 Relevance Vector Machines (RVMs) \citep{BishopTipping00}, by setting $\bphi(\x) = \diag(\balpha^{1/2})\k_{m}(\x)$, where $\balpha$ is an $m \times 1$  vector. Note that to ensure $\K_{nn} - \bPhi\bPhi^\top \succeq \0$, we have to add some constraint over the range of each $\alpha_i$ in $\balpha$. 

The second major advantage of \ours is that our variational ELBO is consistent with the composite non-convex loss form favored by \ps, therefore we can utilize the highly efficient, distributed asynchronous proximal gradient descent in \ps to scale up GPs to extremely large applications (see Section $6.3$). Furthermore, the  simple element-wise and closed-form proximal operation enables exceedingly efficient and parallelizable variational posterior update on the server side.

\section{Experiments}
\subsection{Predictive Performance}
\begin{figure*}[!ht]
	\begin{center}
		\begin{tabular}{cc}
			\includegraphics[width=0.5\linewidth]{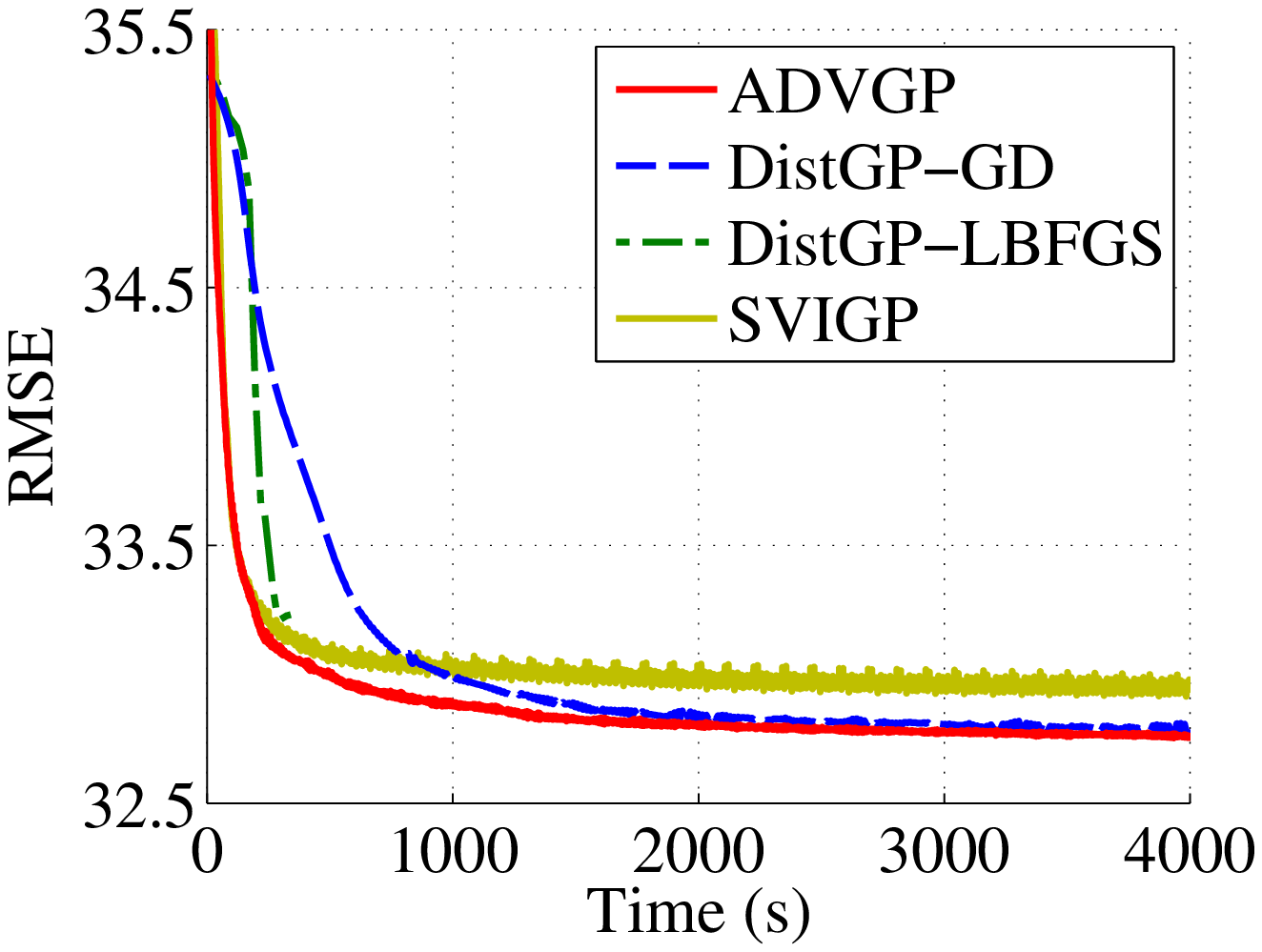} \hspace{-0.03\linewidth} &
			\includegraphics[width=0.5\linewidth]{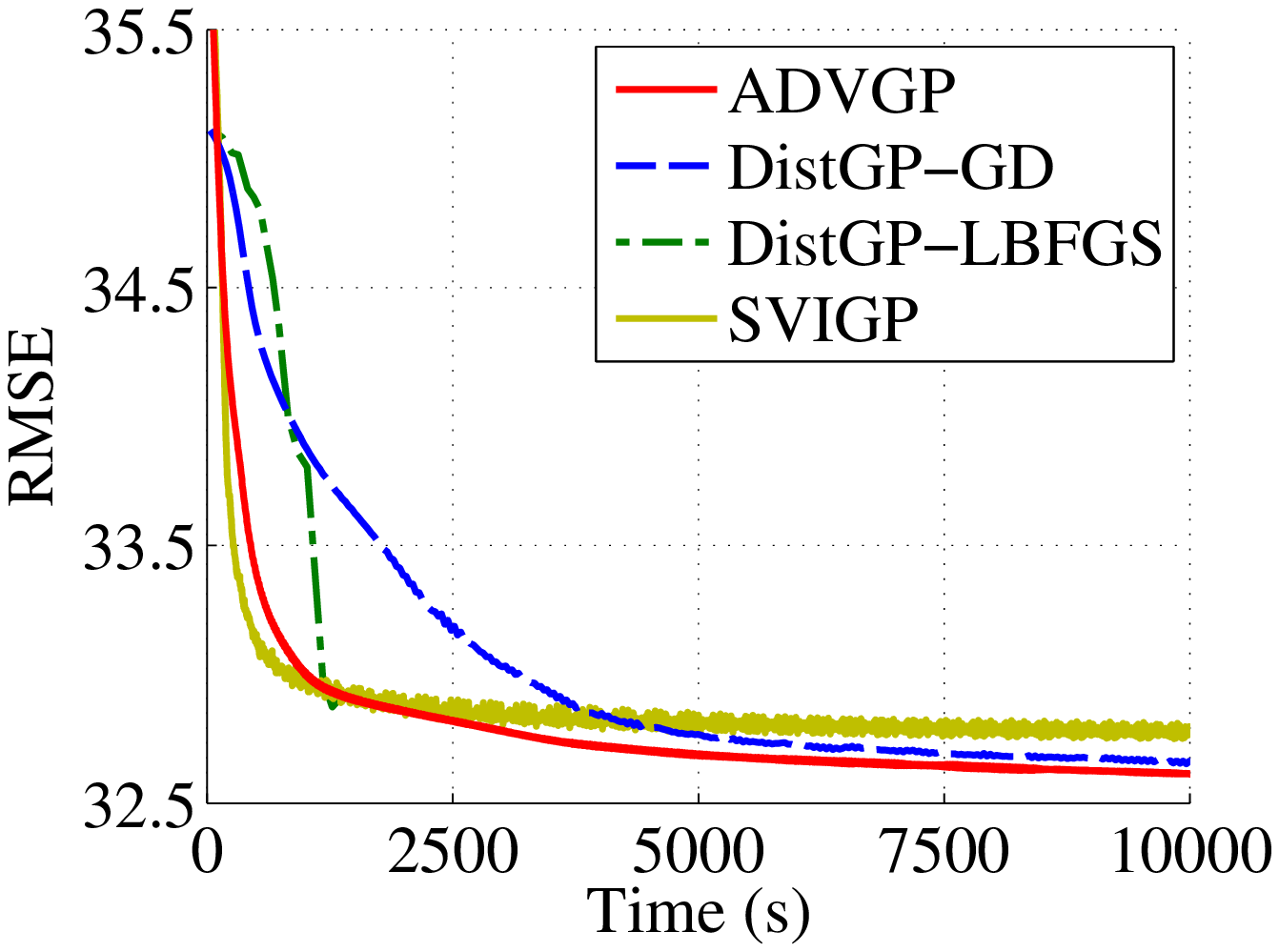} \hspace{-0.03\linewidth} \\
			(A) $n = 700$K, $m = 100$ \hspace{-0.03\linewidth} & 
			(B) $n = 700$K, $m = 200$ \hspace{-0.03\linewidth} \\
			\includegraphics[width=0.5\linewidth]{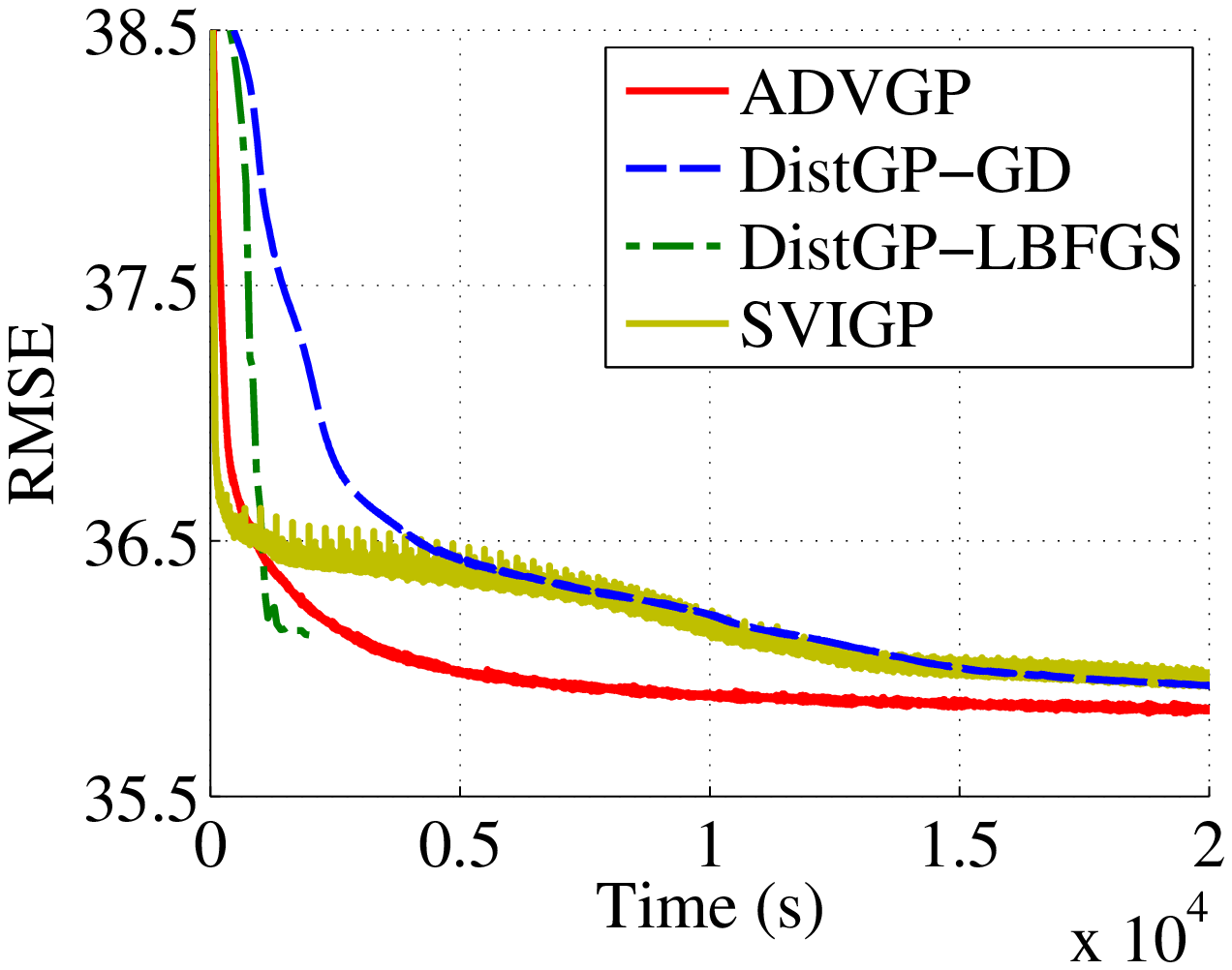} \hspace{-0.03\linewidth} & 
			\includegraphics[width=0.5\linewidth]{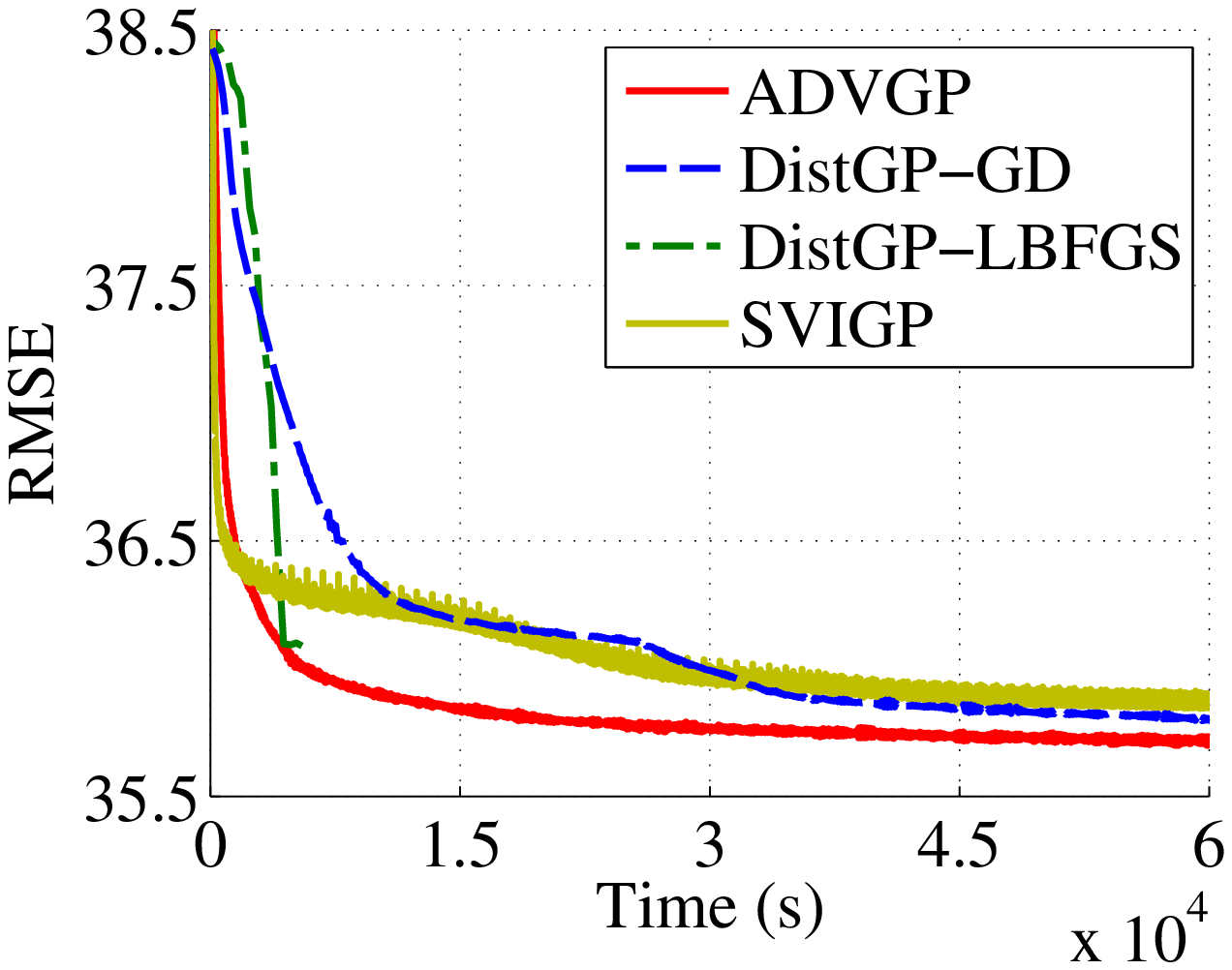} \\
			(C)  $n = 2$M, $m = 100$ \hspace{-0.03\linewidth} &
			(D) $n = 2$M, $m = 200$
		\end{tabular}
	\end{center}
	\caption{Root mean square errors for US flight data as a function of training time.} \label{fig:rmse}
\end{figure*}

First, we evaluated the inference quality of \ours in terms of predictive performance. To this end, we used the US Flight data\footnote{\url{http://stat-computing.org/dataexpo/2009/}} \citep{GPSVI13}, which recorded the  arrival and departure time of the USA commercial flights between January and April in 2008.  We performed two groups of tests:  in the first group, we randomly chose $700$K samples for training; in the second group, we randomly selected $2$M  training samples. Both groups used $100$K samples for testing. We ensured that  the  training and testing data are non-overlapping.  

We compared ADVGP with two existing scalable variational inference algorithms: SVIGP~\citep{GPSVI13} and DistGP~\citep{gal2014distributed}. SVIGP employs an online training,  and DistGP performs a  distributed synchronous variational inference. We ran all the methods on a computer node with $16$  CPU cores and $64$ GB memory. While SVIGP uses a single CPU core, DistGP and \ours use all the CPU cores to perform parallel inference. We used ARD kernel for all the methods, with the same initialization of the kernel parameters. For SVIGP, we set the  mini-batch size to $5000$,  consistent with \citep{GPSVI13}.  For DistGP, we tested two optimization frameworks: local gradient descent (DistGP-GD) and L-BFGS (DistGP-LBFGS). For \ours, we initialized $\bmu = \0, \U = \I$, and used ADADELTA \citep{zeiler12} to adjust the  step size for the gradient descent before the proximal operation. To choose an appropriate delay $\tau$, we  sampled another set of training and test data, based on which we tuned  $\tau$ from $ \{0, 8, 16, 24, 32, 40\}$. These tunning datasets do not overlap the test data in the evaluation. Note that when $\tau=0$, the computation is totally synchronous; larger $\tau$ results in more asynchronous computation. We chose $\tau = 32$ as it produced the best performance on the tunning datasets. 

Table \ref{tab:rmse_700K} and Table \ref{tab:rmse_2M}  report the root mean square errors (RMSEs) of all the methods using different numbers of inducing points, \ie $m \in \{50, 100, 200\}$. As we can see, \ours exhibits better or comparable prediction accuracy in all the cases. Therefore, while using asynchronous computation, \ours maintains the same robustness and quality for inference. Furthermore, we examined the prediction accuracy of each method along with the training time, under the settings $m \in \{100,200\}$. Figure \ref{fig:rmse} shows that during the same time span, \ours achieves the highest performance boost  (\ie RMSE is reduced faster  than the competing  methods), which demonstrates the efficiency of \ours. It is interesting to see that in a short period of time since the beginning, SVIGP  reduces RMSE as fast as \ours; however, after that, RMSE of SVIGP is constantly larger than \ours, exhibiting  an inferior performance. In addition,  DistGP-LBFGS  converges earlier than both \ours and SVIGP. However, RMSE of DistGP-LBFGS  is larger than both \ours and SVIGP at convergence. This implies that the L-BFGS optimization converged  to a suboptimal solution. 

\begin{table}[ht!]
\centering
\begin{subtable}
\centering
\caption{Root mean square errors (RMSEs) for $700$K/$100$K US Flight data.} \label{tab:rmse_700K}
\begin{tabular}{c || c  | c | c}
Method   & $m = 50$ & $m = 100$ & $m = 200$ \\
\hline
Prox GP        & $\bf {32.9080}$ & $\bf {32.7543}$ & $\bf {32.6143}$ \\
GD Dist GP   & $32.9411$ & $32.8069$ & $32.6521$ \\
LBFG Dist GP& $33.0707$ & $33.2263$ & $32.8729$ \\
SVIGP           & $33.1054$ & $32.9499$ & $32.7802$ \\
\end{tabular}
\end{subtable}
\begin{subtable}
\centering
\caption{RMSEs for $2$M/$100$K US Flight data.} \label{tab:rmse_2M}
\begin{tabular}{c || c  | c | c}
Method   & $m = 50$ & $m = 100$ & $m = 200$ \\
\hline
Prox GP         & $36.1156$ & $\bf{35.8347}$ & $\bf {35.7017}$ \\
GD Dist GP    & $36.0142$ & $35.9487 $ & $35.7971$ \\
LBFG Dist GP & $\bf{35.9809}$ & $36.1676$  & $36.0749$ \\
SVIGP            & $36.2019$ & $35.9517$ & $35.8599$ \\
\end{tabular}
\end{subtable}
\end{table}

\begin{figure}[ht!]
\begin{center}
\includegraphics[width=0.8\linewidth]{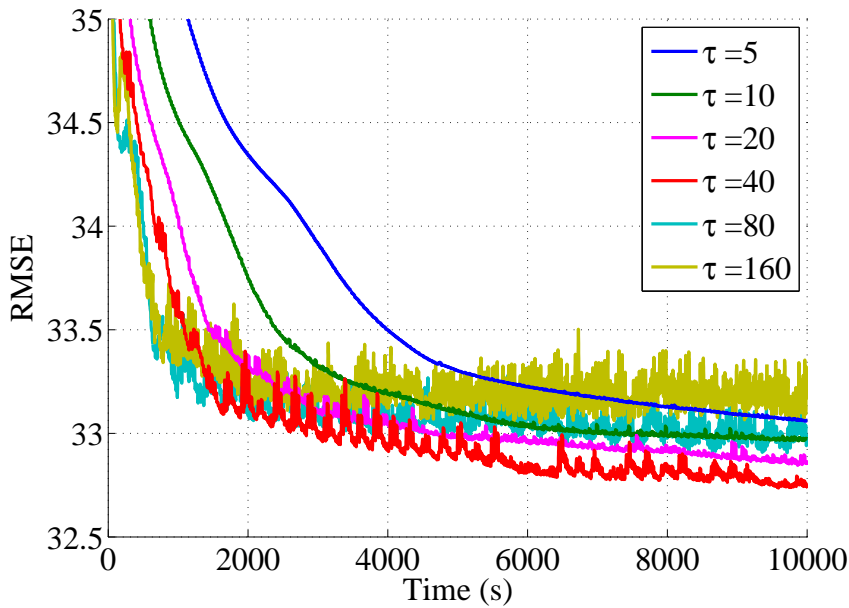}  
\end{center}
\caption{Root mean square errors (RMSEs) as a function of time for different delay limits $\tau$ } \label{fig:diff_delay}
\end{figure}

We also studied how the delay limit $\tau$ affects the performance of \ours. Practically, when many machines are used, some worker may always be slower than the others due to environmental factors, \eg unbalanced workloads. To simulate this scenario, we intentionally introduced a latency by assigning each worker a random sleep time of $0$, $10$ or $20$ seconds at initialization; hence a worker would pause for its given sleep time before each iteration. In our experiment, the average per-iteration running time  was only $0.176$ seconds; so the fastest worker could be hundreds of iterations ahead of the slowest one in the asynchronous setting. We examined $\tau = 0, 5, 10, 20, 40, 80, 160$ and plotted RMSEs as a function of time in Figure \ref{fig:diff_delay}. Since RMSE of the synchronous case ($\tau = 0$) is much larger than the others, we do not show it in the figure. When $\tau$ is larger, \ours's performance is more fluctuating.  
Increasing $\tau$, we first improved the prediction accuracy due to more efficient CPU usage; however, later we observed a decline caused by the excessive asynchronization that impaired the optimization. Therefore, to use \ours for workers at various paces, we need to carefully choose the appropriate delay limit $\tau$.

\subsection{Scalability}
\begin{figure}[ht!]
\vspace{-0.1in}
\begin{center}
\begin{tabular}{cc}
\includegraphics[width=0.5\linewidth]{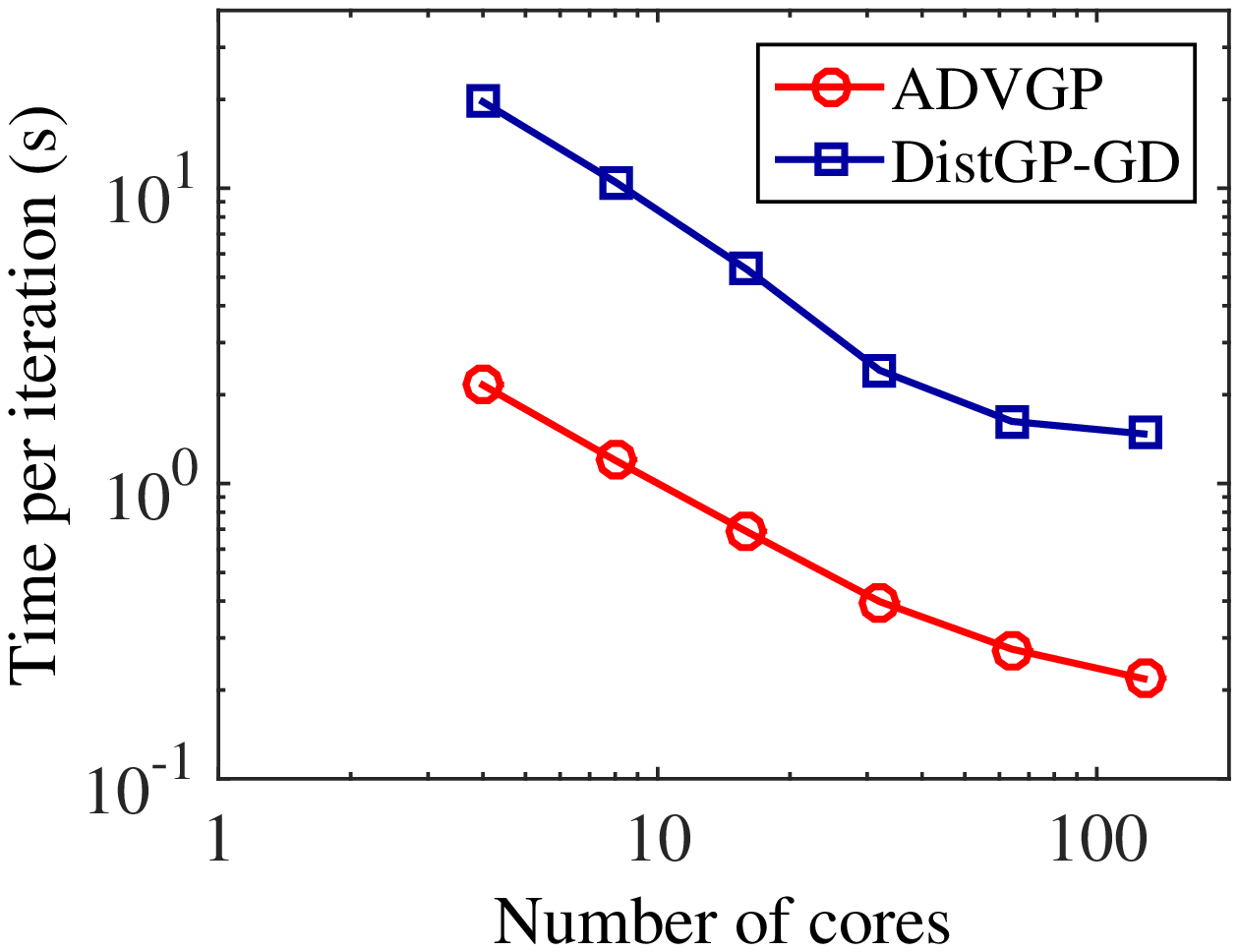}   \hspace{-0.05\linewidth} &
  \includegraphics[width=0.5\linewidth]{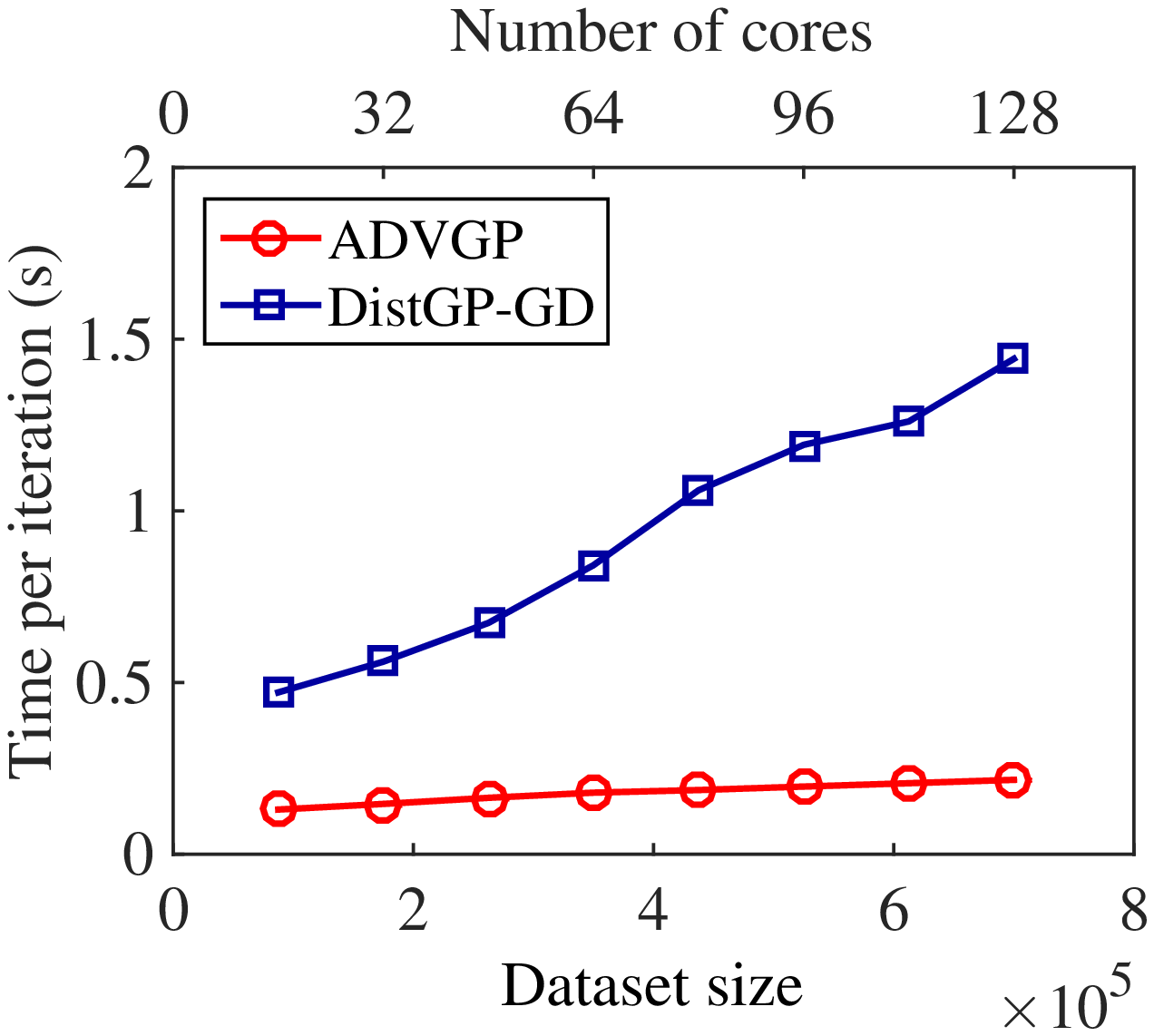}\\
(A)  \hspace{-0.05\linewidth} & (B)
\end{tabular}
\end{center}
\caption{Scalability tests on $700$K US flight data. (A) Per-iteration time as a function of available cores in log-scale. (B) Per-iteration time when scaling the computational resources proportionally to dataset size.} \label{fig:scalability}
\end{figure}

Next, we examined the scalability of our asynchronous inference method, \ours. To this end, we used the $700$K/$100$K  dataset and compared with the synchronous inference algorithm DistGP~\citep{gal2014distributed}. For a fair comparison, we used the local gradient descent version of DistGP, \ie DistGP-GD. We conducted two experiments on 4 c4.8xlarge instances of Amazon EC2 cloud,  where we set the number of inducing points $m = 100$. In the first experiment, we fixed the size of the training data, and increased the number of  CPU cores from $4$ to $128$. We examined the per-iteration running time of both \ours and DistGP-GD. Figure \ref{fig:scalability}(A) shows that while both decreasing with more CPU cores, the per-iteration running time of \ours is much less than that of DistGP-GD. This demonstrates the advantage of \ours in computational efficiency. In addition, the per-iteration running time of \ours decays much more quickly than that of DistGP-GD as the number of cores approaches $128$. 
This implies that even the communication cost becomes dominant, the asynchronous mechanism of \ours still effectively reduces the latency and maintains a high usage of the computational power. 
In the second experiment, we simultaneously increased the number of cores and the size of training data. We started from $87.5$K samples and $16$ cores and gradually increased  them to $700$K samples and $128$ cores.  As shown in Figure \ref{fig:scalability}(B), the average per-iteration time of DistGP-GD grows linearly; in contrast, the average per-iteration time of \ours stays almost constant. We speculate that without synchronous coordination, \ours can fully utilize the network bandwidth so that the increased amount of messages, along with the growth of the data size, affect little the network communication efficiency. This demonstrates the advantage of asynchronous inference from another perspective.

\vspace{-0.1in}
\subsection{NYC Taxi Traveling Time Prediction}
Finally, we applied \ours for an  extremely large problem: the prediction of the taxi traveling time in New York city. We used  the  New York city yellow taxi trip 
dataset  \footnote{\url{http://www.nyc.gov/html/tlc/html/about/trip_record_data.shtml}}, which consist of $1.21$ billions of trip records from January 2009 to December 2015.  We excluded the trips  that are  outside the NYC area or more than 5 hours. The average traveling time is 764 seconds and the standard derivation is  576 seconds. To predict the traveling time, we used the following 9 features: time of the day, day of the week, day of the month, month, pick-up latitude, pick-up longitude, drop-off latitude, drop-off longitude, and travel distance. We used Amazon EC2 cloud, and ran \ours on multiple Amazon c4.8xlarge instances, each with 36 vCPUs and 60 GB memory. We compared with the  linear regression model implemented in Vowpal Wabbit~\citep{agarwal14}. Vowpal Wabbit is a state-of-the-art large scale machine learning software package and has been used in many industrial-scale applications, such as click-through-rate prediction~\citep{olivier2014simple}. 

\begin{figure}[ht!]
\vspace{-0.1in}
\begin{center}
\begin{tabular}{cc}
\includegraphics[width=0.5\linewidth]{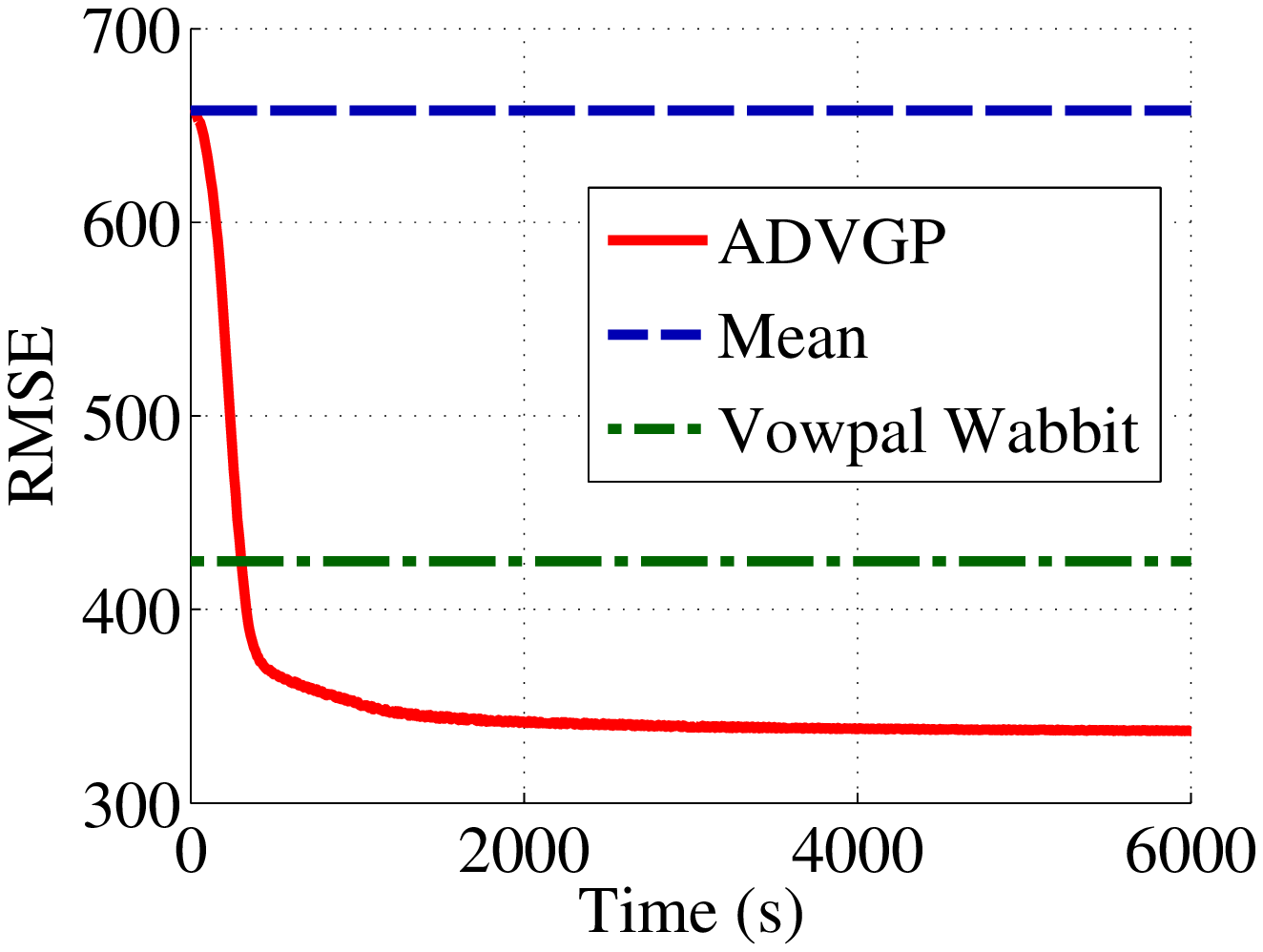} \hspace{-0.05\linewidth} & \includegraphics[width=0.5\linewidth]{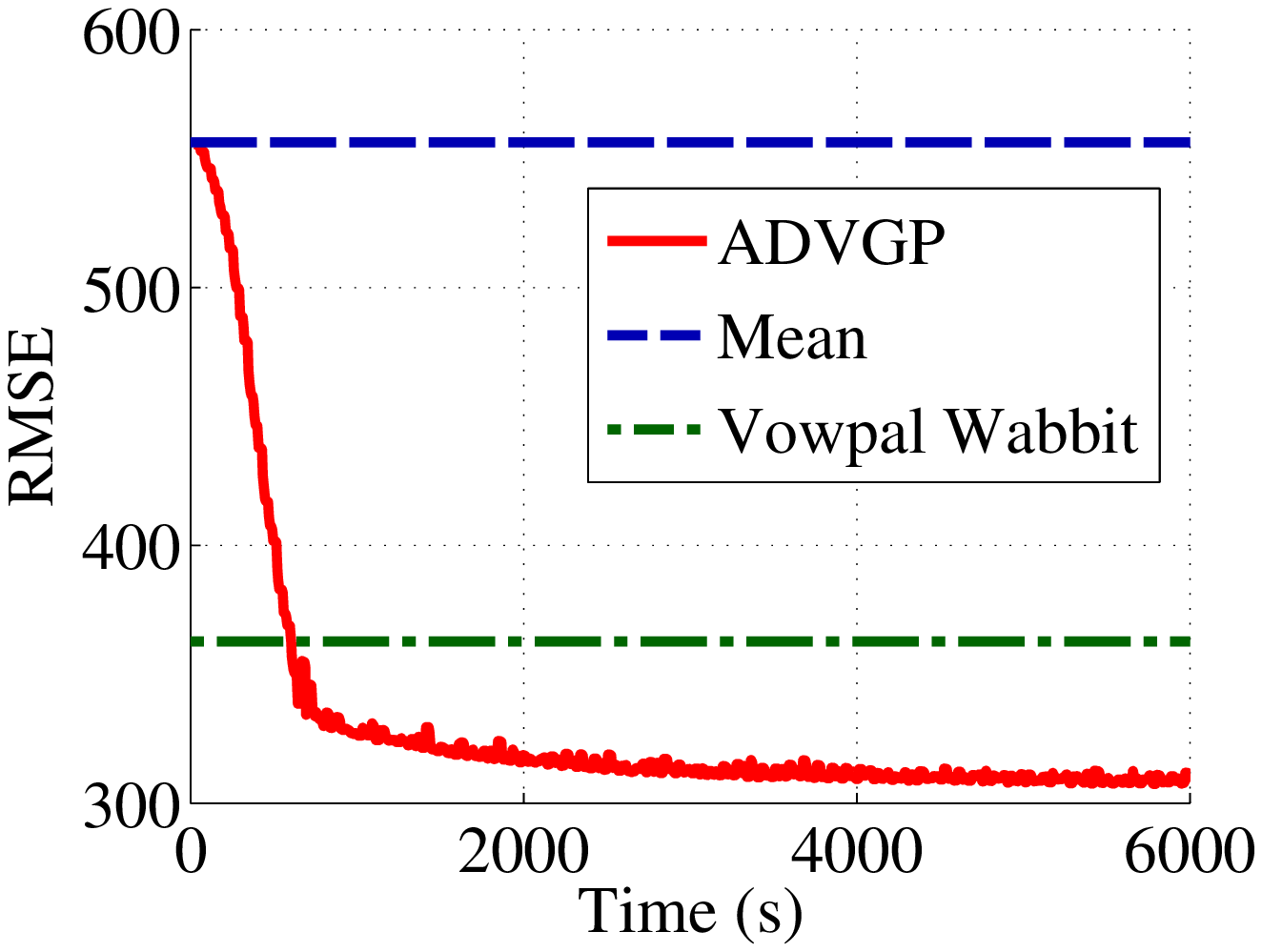}\\
(A) 100M training samples \hspace{-0.05\linewidth}  & (B) 1B training samples
\end{tabular}
\end{center}
\caption{RMSE as a function of training time on NYC Taxi Data.} \label{fig:taxi_rmse}
\end{figure}

We first randomly selected $100$M training samples and $500$K test samples. We set $m=50$ and initialized the inducing points as the 
the K-means cluster centers from a subset of $2$M training samples. We trained a GP regression model with \ours, using $5$ Amazon instances with $200$ processes. The delay limit $\tau$ was selected as $20$. We used Vowpal Wabbit  to train a linear regression model, with default settings. We also  took the average traveling time over the training data to obtain a simple mean prediction. In Figure \ref{fig:taxi_rmse}(A), we report RMSEs of the linear regression and the mean prediction, as well as the GP regression along with running time.  As we can see,  \ours greatly outperforms the competing methods. Only after $6$ minutes, \ours has improved RMSEs  of the linear regression and the mean prediction by $9\%$ and $41\%$, respectively; the improvements continued for about $30$ minutes. Finally, \ours reduced the RMSEs of the linear regression and the mean prediction by  $22\%$ and $49\%$, respectively. The RMSEs are \{\ours: $333.4$, linear regression: $424.8$,  mean-prediction: $657.7$\}.


To further verify the advantage of GP regression in extremely large applications, we  used $1$B  training and $1$M testing samples. We used $50$ inducing points, initialized by the K-means cluster centers from a $1$M training subset. We ran \ours using $28$ Amazon instances with $1000$ processes and chose $\tau=100$. As shown in Figure \ref{fig:taxi_rmse}(B),  the RMSE of GP regression outperforms the linear models by a large margin. 
After $12$ minutes, \ours has improved the RMSEs of the linear regression and the mean prediction by $9\%$ and $66\%$, respectively; the improvement kept growing for about $1.5$ hours.  At the convergence, \ours outperforms the linear regression and the mean prediction by $17\%$ and $80\%$, respectively. The RMSEs are \{\ours: $309.7$, linear regression: $362.8$,  mean-prediction: $556.3$\}. In addition, the average per-iteration time of \ours is only $0.21$ seconds.
These results confirm the power of the nonlinear regression in extremely large real-world scenarios, comparing with linear models, while the latter are much easier to be scaled up and hence more popular.  

\vspace{-0.1in}
\section{Conclusion}
We have presented \ours, an asynchronous, distributed variational inference algorithm for GP regression,  which enables real-world extremely large applications. \ours is based on a novel variational GP framework, which allows flexible construction of low rank approximations and can relate to many sparse GP models.

\bibliography{adgp}

\begin{thebibliography}{}

\bibitem[Agarwal and Duchi(2011)Agarwal and Duchi]{agarwal2011distributed}
Agarwal, A. and Duchi, J.~C. (2011).
\newblock Distributed delayed stochastic optimization.
\newblock In {\em Advances in Neural Information Processing Systems 24\/},
  pages 873--881. Curran Associates, Inc.

\bibitem[Agarwal {\em et~al.}(2014)Agarwal, Chapelle, Dud\'{i}k, and
  Langford]{agarwal14}
Agarwal, A., Chapelle, O., Dud\'{i}k, M., and Langford, J. (2014).
\newblock A reliable effective terascale linear learning system.
\newblock {\em Journal of Machine Learning Research\/}, {\bf 15}, 1111--1133.

\bibitem[Bauer {\em et~al.}(2016)Bauer, van~der Wilk, and
  Rasmussen]{bauer2016understanding}
Bauer, M., van~der Wilk, M., and Rasmussen, C.~E. (2016).
\newblock Understanding probabilistic sparse {G}aussian process approximations.
\newblock In {\em Advances in Neural Information Processing Systems 29\/},
  pages 1525--1533.

\bibitem[Bishop and Tipping(2000)Bishop and Tipping]{BishopTipping00}
Bishop, C.~M. and Tipping, M.~E. (2000).
\newblock Variational relevance vector machines.
\newblock In {\em Proceedings of the 16th Conference in Uncertainty in
  Artificial Intelligence (UAI)\/}.

\bibitem[Chapelle {\em et~al.}(2014)Chapelle, Manavoglu, and
  Rosales]{olivier2014simple}
Chapelle, O., Manavoglu, E., and Rosales, R. (2014).
\newblock Simple and scalable response prediction for display advertising.
\newblock {\em ACM Transactions on Intelligent Systems and Technology
  (TIST)\/}, {\bf 5}(4), 61:1--61:34.

\bibitem[Dai {\em et~al.}(2014)Dai, Damianou, Hensman, and
  Lawrence]{dai2014gaussian}
Dai, Z., Damianou, A., Hensman, J., and Lawrence, N.~D. (2014).
\newblock {G}aussian process models with parallelization and {GPU}
  acceleration.
\newblock In {\em NIPS Workshop on Software Engineering for Machine
  Learning\/}.

\bibitem[Deisenroth and Ng(2015)Deisenroth and Ng]{deisenroth2015distributed}
Deisenroth, M. and Ng, J.~W. (2015).
\newblock Distributed {G}aussian processes.
\newblock In {\em Proceedings of the 32nd International Conference on Machine
  Learning\/}, pages 1481--1490.

\bibitem[Filippone {\em et~al.}(2013)Filippone, Zhong, and
  Girolami]{filippone2013comparative}
Filippone, M., Zhong, M., and Girolami, M. (2013).
\newblock A comparative evaluation of stochastic-based inference methods for
  gaussian process models.
\newblock {\em Machine Learning\/}, {\bf 93}(1), 93--114.

\bibitem[Gal {\em et~al.}(2014)Gal, van~der Wilk, and
  Rasmussen]{gal2014distributed}
Gal, Y., van~der Wilk, M., and Rasmussen, C. (2014).
\newblock Distributed variational inference in sparse {G}aussian process
  regression and latent variable models.
\newblock In {\em Advances in Neural Information Processing Systems 27\/},
  pages 3257--3265.

\bibitem[Hensman {\em et~al.}(2013)Hensman, Fusi, and Lawrence]{GPSVI13}
Hensman, J., Fusi, N., and Lawrence, N.~D. (2013).
\newblock Gaussian processes for big data.
\newblock In {\em Proceedings of the Conference on Uncertainty in Artificial
  Intelligence (UAI)\/}.

\bibitem[Hensman {\em et~al.}(2015)Hensman, Matthews, Filippone, and
  Ghahramani]{hensman2015mcmc}
Hensman, J., Matthews, A.~G., Filippone, M., and Ghahramani, Z. (2015).
\newblock Mcmc for variationally sparse gaussian processes.
\newblock In {\em Advances in Neural Information Processing Systems\/}, pages
  1648--1656.

\bibitem[Kumar {\em et~al.}(2009)Kumar, Mohri, and
  Talwalkar]{kumar2009ensemble}
Kumar, S., Mohri, M., and Talwalkar, A. (2009).
\newblock Ensemble {N}ystr\"om method.
\newblock In Y.~Bengio, D.~Schuurmans, J.~D. Lafferty, C.~K.~I. Williams, and
  A.~Culotta, editors, {\em Advances in Neural Information Processing Systems
  22\/}, pages 1060--1068. Curran Associates, Inc.

\bibitem[Li {\em et~al.}(2013)Li, Andersen, and Smola]{li2013distributed}
Li, M., Andersen, D.~G., and Smola, A.~J. (2013).
\newblock Distributed delayed proximal gradient methods.
\newblock In {\em NIPS Workshop on Optimization for Machine Learning\/}.

\bibitem[Li {\em et~al.}(2014a)Li, Andersen, Smola, and
  Yu]{li2014communication}
Li, M., Andersen, D.~G., Smola, A., and Yu, K. (2014a).
\newblock Communication efficient distributed machine learning with the
  parameter server.
\newblock In {\em Neural Information Processing Systems 27\/}.

\bibitem[Li {\em et~al.}(2014b)Li, Andersen, Park, Smola, Ahmed, Josifovski,
  Long, Shekita, and Su]{li2014scaling}
Li, M., Andersen, D.~G., Park, J.~W., Smola, A.~J., Ahmed, A., Josifovski, V.,
  Long, J., Shekita, E.~J., and Su, B.-Y. (2014b).
\newblock Scaling distributed machine learning with the parameter server.
\newblock In {\em 11th USENIX Symposium on Operating Systems Design and
  Implementation (OSDI 14)\/}, pages 583--598.

\bibitem[Matthews {\em et~al.}(2017)Matthews, van~der Wilk, Nickson, Fujii,
  Boukouvalas, Le{\'o}n-Villagr{\'a}, Ghahramani, and
  Hensman]{matthews2017gpflow}
Matthews, A. G. d.~G., van~der Wilk, M., Nickson, T., Fujii, K., Boukouvalas,
  A., Le{\'o}n-Villagr{\'a}, P., Ghahramani, Z., and Hensman, J. (2017).
\newblock {GPflow}: A {G}aussian process library using {T}ensor{F}low.
\newblock {\em Journal of Machine Learning Research\/}, {\bf 18}(40), 1--6.

\bibitem[Murray and Adams(2010)Murray and Adams]{murray2010slice}
Murray, I. and Adams, R.~P. (2010).
\newblock Slice sampling covariance hyperparameters of latent gaussian models.
\newblock In {\em Advances in Neural Information Processing Systems 24\/},
  pages 1732--1740.

\bibitem[Peng and Qi(2015)Peng and Qi]{hao2015eigengp}
Peng, H. and Qi, Y. (2015).
\newblock {EigenGP}: Sparse {G}aussian process models with adaptive
  eigenfunctions.
\newblock In {\em Proceedings of the 24th International Joint Conference on
  Artificial Intelligence\/}, pages 3763--3769.

\bibitem[Qui{\~n}onero-Candela and Rasmussen(2005)Qui{\~n}onero-Candela and
  Rasmussen]{quinonero2005unifying}
Qui{\~n}onero-Candela, J. and Rasmussen, C.~E. (2005).
\newblock A unifying view of sparse approximate {G}aussian process regression.
\newblock {\em The Journal of Machine Learning Research\/}, {\bf 6},
  1939--1959.

\bibitem[Rasmussen and Williams(2006)Rasmussen and Williams]{GPML06}
Rasmussen, C.~E. and Williams, C. K.~I. (2006).
\newblock {\em Gaussian Processes for Machine Learning\/}.
\newblock The MIT Press.

\bibitem[Schwaighofer and Tresp(2003)Schwaighofer and
  Tresp]{schwaighofer2003transductive}
Schwaighofer, A. and Tresp, V. (2003).
\newblock Transductive and inductive methods for approximate {G}aussian process
  regression.
\newblock In {\em Advances in Neural Information Processing Systems 15\/},
  pages 953--960. MIT Press.

\bibitem[Seeger {\em et~al.}(2003)Seeger, Williams, and
  Lawrence]{seeger2003fast}
Seeger, M., Williams, C., and Lawrence, N. (2003).
\newblock Fast forward selection to speed up sparse {G}aussian process
  regression.
\newblock In {\em Proceedings of the Ninth International Workshop on Artificial
  Intelligence and Statistics\/}.

\bibitem[Smola and Bartlett(2001)Smola and Bartlett]{smola2001sparse}
Smola, A.~J. and Bartlett, P.~L. (2001).
\newblock Sparse greedy {G}aussian process regression.
\newblock In {\em Advances in Neural Information Processing Systems 13\/}. MIT
  Press.

\bibitem[Snelson and Ghahramani(2005)Snelson and Ghahramani]{snelson2005sparse}
Snelson, E. and Ghahramani, Z. (2005).
\newblock Sparse {G}aussian processes using pseudo-inputs.
\newblock In {\em Advances in Neural Information Processing Systems\/}, pages
  1257--1264.

\bibitem[Titsias(2009)Titsias]{titsias2009variational}
Titsias, M.~K. (2009).
\newblock Variational learning of inducing variables in sparse {G}aussian
  processes.
\newblock In {\em Proceedings of the Twelfth International Conference on
  Artificial Intelligence and Statistics\/}, pages 567--574.

\bibitem[Williams and Seeger(2001)Williams and Seeger]{williams2001using}
Williams, C. and Seeger, M. (2001).
\newblock Using the {N}ystr{\"o}m method to speed up kernel machines.
\newblock In {\em Advances in Neural Information Processing Systems 13\/},
  pages 682--688.

\bibitem[Zeiler(2012)Zeiler]{zeiler12}
Zeiler, M.~D. (2012).
\newblock {ADADELTA:} an adaptive learning rate method.
\newblock {\em arXiv:1212.5701\/}.

\end{thebibliography}
\bibliographystyle{natbib}

\appendix
\part*{Appendices}

\counterwithin{figure}{section}
\counterwithin{table}{section}
\section{Derivatives}
\subsection{Objective}
As described in the paper, the objective function to be minimized is $-\mathcal{L} = \sum_{i=1}^{n} g_i + h,$ where
\begin{align}
g_i & = -\ln \mathcal{N}(y_n | \boldsymbol{\phi}_i^{\mathrm{T}}\boldsymbol{\mu}, \beta^{-1})+\frac{\beta}{2} \boldsymbol{\phi}_i^{\mathrm{T}}\boldsymbol{\Sigma} \boldsymbol{\phi}_i+\frac{\beta}{2}\tilde{k}_{ii} \notag\\
& = \frac{1}{2} \ln 2 \pi  -\frac{1}{2} \ln \beta + \frac{\beta}{2} \left(y_i^2 - 2 y_i \boldsymbol{\phi}_i^{\mathrm{T}} \boldsymbol{\mu} + \boldsymbol{\mu}^{\mathrm{T}} \boldsymbol{\phi}_i\boldsymbol{\phi}_i^{\mathrm{T}}\boldsymbol{\mu} + \boldsymbol{\phi}_i^{\mathrm{T}} \boldsymbol{\Sigma} \boldsymbol{\phi}_i + k_{ii} - \boldsymbol{\phi}_i^{\mathrm{T}}\boldsymbol{\phi}_i \right),  \label{eq:g} \\
h & = \mathrm{KL}(q(\boldsymbol w)  || p(\boldsymbol w) ) \notag\\
& =\frac{1}{2} \left(-\ln |\boldsymbol{\Sigma}| - m + \mathrm{tr}(\boldsymbol{\Sigma}) + \boldsymbol{\mu}^{\mathrm{T}}\boldsymbol{\mu} \right), \label{eq:h}
\end{align}
and we define $\beta = \sigma^{-2}$ and $\boldsymbol{\phi}_i = \boldsymbol{\phi}(\boldsymbol{x}_i)$.

\subsection{Kernel}
A common choice for the kernel is the anisotropic squared exponential covariance function:
\begin{align}
k(\boldsymbol{x}_i, \boldsymbol{x}_j) = a_0^2 \exp\big(- \frac{1}{2}(\boldsymbol{x}_i-\boldsymbol{x}_j)^{T} \mathrm{diag}(\boldsymbol{\eta}) (\boldsymbol{x}_i-\boldsymbol{x}_j)\big),
\end{align}
in which the hyperparameters are the signal variance  $a_0$ and the lengthscales $\boldsymbol{\eta}=\{1/a_k^2\}_{k=1}^d$, controlling how fast the covariance decays with the distance between inputs.
Using this covariance function, we can prune input dimensions by shrinking the corresponding lengthscales based on the data (when $\eta_d=0$, the $d$-th dimension becomes totally irrelevant to the covariance function value). This pruning is known as Automatic Relevance Determination (ARD) and therefore this covariance is also called the ARD squared exponential.

{\bf Derivative over $\ln \sigma$}

The derivative of $g_{i}$ over $\ln \sigma$ is
\begin{align}
\frac{\partial g_i}{\partial \ln \sigma} = 1 -  \frac{1}{\sigma^2} (y_i^2 - 2 y_i \boldsymbol{\phi}_i^{T} \boldsymbol{\mu} + \boldsymbol{\phi}_i^{T}(\boldsymbol{\Sigma} + \boldsymbol{\mu}\boldsymbol{\mu}^T) \boldsymbol{\phi}_i + k_{ii} - \boldsymbol{\phi}_i^T\boldsymbol{\phi}_i).
\end{align}

{\bf  Derivative over $\ln a_0$ }

The derivative of $g_{i}$ over $\ln a_{0}$ is
\begin{align}
\frac{\partial g_i}{\partial \ln a_0 } = \frac{1}{\sigma^2} (-y_i \boldsymbol{\phi}_i^{T} \boldsymbol{\mu} + \boldsymbol{\phi}_i^{T}(\boldsymbol{\Sigma} + \boldsymbol{\mu}\boldsymbol{\mu}^T) \boldsymbol{\phi}_i + k_{ii} - \boldsymbol{\phi}_i^T\boldsymbol{\phi}_i).
\end{align}

{\bf Derivative over $\boldsymbol{Z}$ }

By defining $\boldsymbol{L}$ the lower triangular Cholesky factor of $\boldsymbol{K}_{mm}^{-1}$, the derivative of $g_{i}$ over $\boldsymbol{Z}$ is
\begin{align}
\frac{\partial g_i}{\partial \boldsymbol{Z }} = & \frac{1}{\sigma^2} \big[ \left( (\boldsymbol{L} \boldsymbol{p}_i) \circ \boldsymbol{k}_m(\boldsymbol{x}_i) \right) \boldsymbol{x}_i^T \mathrm{diag}(\eta) - \left( ( (\boldsymbol{L} \boldsymbol{p}_i) \circ \boldsymbol{k}_m(\boldsymbol{x}_i)) \boldsymbol{1}_d^T \right) \circ (\boldsymbol{Z} \mathrm{diag}(\boldsymbol{\eta})  ) \nonumber \\
&  - (\boldsymbol{T}_i + \boldsymbol{T}_i^{T}) \boldsymbol{Z} \mathrm{diag}(\boldsymbol{\eta})  + ((\boldsymbol{T}_i + \boldsymbol{T}_i^{T}) \boldsymbol{1}_m \boldsymbol{\eta}^T)\circ \boldsymbol{Z} \big],
\end{align}
where
\begin{align}
\boldsymbol{p_i} & = -\boldsymbol{\mu} y_i + (\boldsymbol{\mu} \boldsymbol{\mu}^T +\boldsymbol{\Sigma}) \boldsymbol{\phi}(\boldsymbol{x}_i) - \boldsymbol{\phi}(\boldsymbol{x}_i),\\
\boldsymbol{T_i} &= \left[ \boldsymbol{L} \left( (\boldsymbol{\phi}(\boldsymbol{x}_i) \boldsymbol{p}_i^T) \circ \boldsymbol{\Psi} \right) \boldsymbol{L}^T \right] \circ \boldsymbol{K}_{mm}.
\end{align}
The symbol $\circ$ denotes the Hadamard product, and $\boldsymbol{\Psi}$ is an upper triangular matrix with diagonal elements all equal to $0.5$ and strictly upper triangular elements all equal to $1$, as follows:
\begin{align}
\boldsymbol{\Psi}  = \left[ \begin{array}{ccccc} 0.5 &1 & \hdots &1 & 1\\ 
							0 & 0.5 &\ddots & 1 & 1 \\
							\vdots & \ddots &  \ddots & \ddots & \vdots \\
							0 &  0 &  \ddots & 0.5 & 1 \\ 
							0 &  0 &  \hdots & 0 & 0.5  \end{array}\right].
\end{align}

{\bf Derivative over $\ln \boldsymbol{\eta}$ }

The derivative of $g_{i}$ over $\ln \boldsymbol{\eta}$ is
\begin{align}
\frac{\partial g_i}{\partial \ln \boldsymbol{\eta}} = & \frac{1}{2\sigma^2} \Big\{
2  \boldsymbol{1}_m^{\mathrm{T}} \left[ \boldsymbol{Z} \circ  \left(\left( (\boldsymbol{L} \boldsymbol{p}_i) \circ \boldsymbol{k}_m(\boldsymbol{x}_i)\right) \boldsymbol{x}_i^T \right)  \right]
- \boldsymbol{1}_m^{\mathrm{T}}  \left( (\boldsymbol{L} \boldsymbol{p}_i) \circ \boldsymbol{k}_m(\boldsymbol{x}_i)\right) (\boldsymbol{x}_i  \circ \boldsymbol{x}_i  ) ^{\mathrm{T}} \notag \\
 & -  \left( (\boldsymbol{L} \boldsymbol{p}_i) \circ \boldsymbol{k}_m(\boldsymbol{x}_i)\right)^{\mathrm{T}} (\boldsymbol{Z} \circ \boldsymbol{Z} )  - \boldsymbol{1}_m^{\mathrm{T}}  \left[ \boldsymbol{Z} \circ (( \boldsymbol{T}_i + \boldsymbol{T}_i^{\mathrm{T}})\boldsymbol{Z}) \right]  
+\boldsymbol{1}_m^{\mathrm{T}}  \left[ (\boldsymbol{T}_i + \boldsymbol{T}_i^T)(\boldsymbol{Z} \circ \boldsymbol{Z}) \right]
\Big\} \circ \boldsymbol{\eta}.
\end{align}

\section{Properties of the ELBO of ADVGP}
By defining $\boldsymbol{U}$ as the upper triangular Cholesky factor of $\boldsymbol{\Sigma}$, i.e., $\boldsymbol{\Sigma} = \boldsymbol{U}^{\mathrm{T}} \boldsymbol{U}$, we have
\begin{lemma}
The gradient of $g_i$ in Equation \ref{eq:g}, $\nabla g_i$, is Lipschitz continuous with respect to each element in $\boldsymbol{\mu}$ and $\boldsymbol{U}$.
\end{lemma}
We can prove this by showing the first derivative of $\nabla g_i$  with respect to each element of $\boldsymbol{\mu}$ and $\boldsymbol{U}$ is bounded, which is constant in our case. As shown in our paper, the gradients of $g_i$ with respect to  $\boldsymbol{\mu}$ and $\boldsymbol{U}$ are:
\begin{align}
\frac{\partial g_i}{\partial \boldsymbol{\mu}} &= \frac{1}{\sigma^2} [-y_i \boldsymbol{\phi}_i +\boldsymbol{\phi}_i\boldsymbol{\phi}_i^{T}\boldsymbol{\mu}], \\
\frac{\partial g_i}{\partial \boldsymbol{U}} &= \frac{1}{\sigma^2} \mathrm{triu}[\boldsymbol{U} \boldsymbol{\phi}_i \boldsymbol{\phi}_i^T],
\end{align}
which are affine functions for $\boldsymbol{\mu}$ and $\boldsymbol{\Sigma}$ respectively. Therefore, the first derivative of $\nabla g_i$ is constant.

\begin{lemma}
$h$ in Equation \ref{eq:h} is a convex function with respect to $\boldsymbol{\mu}$ and $\boldsymbol{U}$.
\end{lemma}
This can be proved by verifying that the Hessian matrices of $h$ with respect to $\boldsymbol{\mu}$ and $\mathrm{vec(\boldsymbol{U})}$ are both positive semidefinite, where we denote $\mathrm{vec(\cdot)}$ as the operator that stacks the columns of a matrix as a vector. To show this, we first compute the partial derivatives of $h$ with respect to $\boldsymbol{\mu}$ and $\boldsymbol{U}$ as
\begin{align}
\frac{\partial h}{\partial \boldsymbol{\mu}} &= \boldsymbol{\mu}, \\
\frac{\partial h}{\partial \boldsymbol{U}} &= -\mathrm{diag}(\boldsymbol{U}^{-1}) + \boldsymbol{U}.
\end{align}
The Hessian matrix of $h$ with respect to $\boldsymbol{\mu}$ is
\begin{align}
\boldsymbol{H}(\boldsymbol{\mu}) = \boldsymbol{I}_{m \times m}  \succeq 0.
\end{align}
The Hessian matrix of $h$ with respect to $\mathrm{vec}(\boldsymbol{U})$ is
\begin{align}
\boldsymbol{H}(\mathrm{vec}(\boldsymbol{U})) = \mathrm{diag}(\boldsymbol{h}) \succeq 0,
\end{align}
where $\boldsymbol{h} = [\frac{\partial h}{\partial U_{11}^2}, \ldots,  \frac{\partial h}{\partial U_{1m}^2}, \ldots, \frac{\partial h}{\partial U_{m1}^2}, \ldots,  \frac{\partial h}{\partial U_{mm}^2} ] $, and $\frac{\partial h}{\partial U_{ij}^2} = 1+\delta(i,j)\frac{1}{U^2_{i,i}}$.


\section{Negative Log Evidences on US Flight Data}
\begin{figure}[ht!]
\begin{center}
\begin{tabular}{cc}
\includegraphics[width=0.45\linewidth]{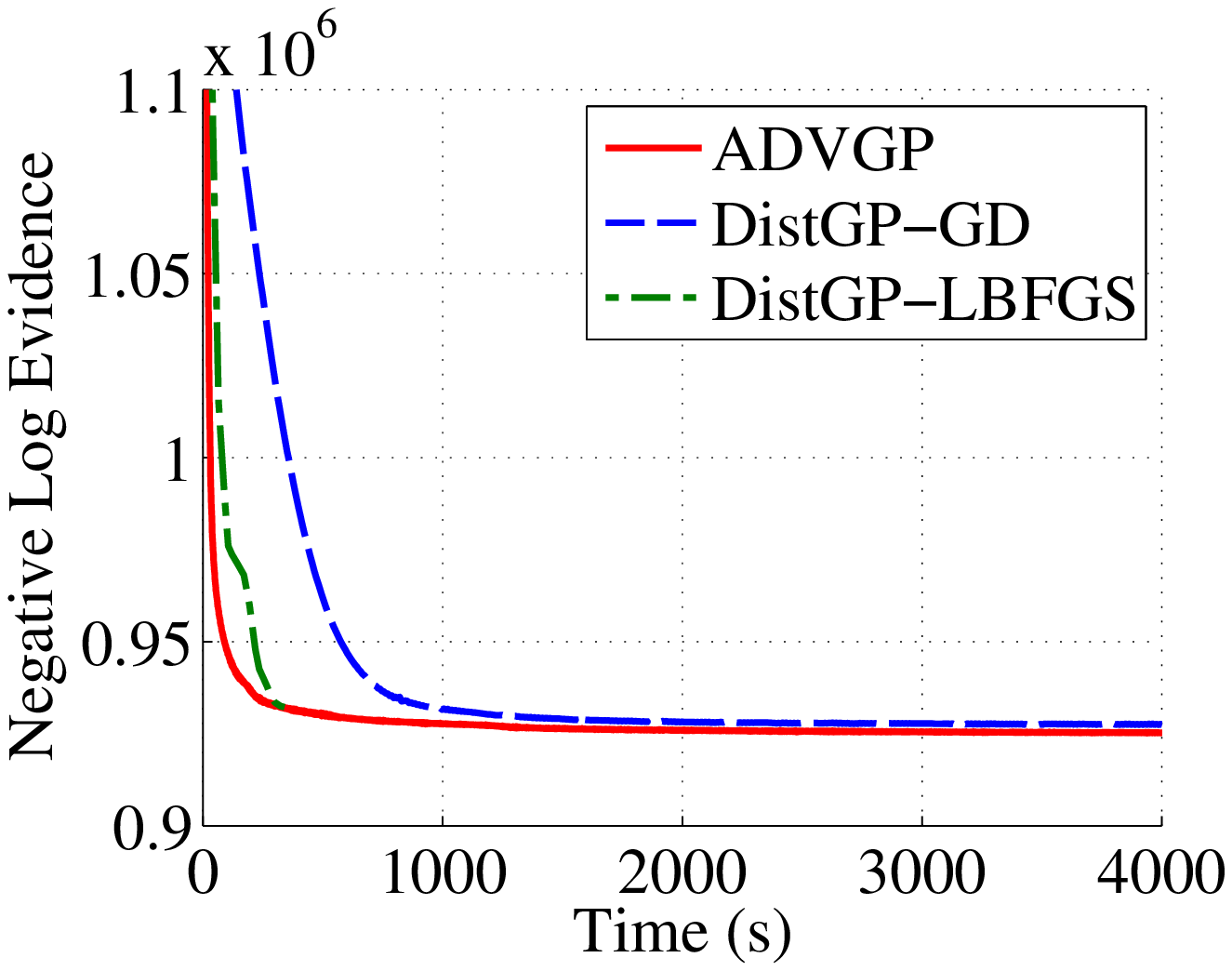}  &
\includegraphics[width=0.45\linewidth]{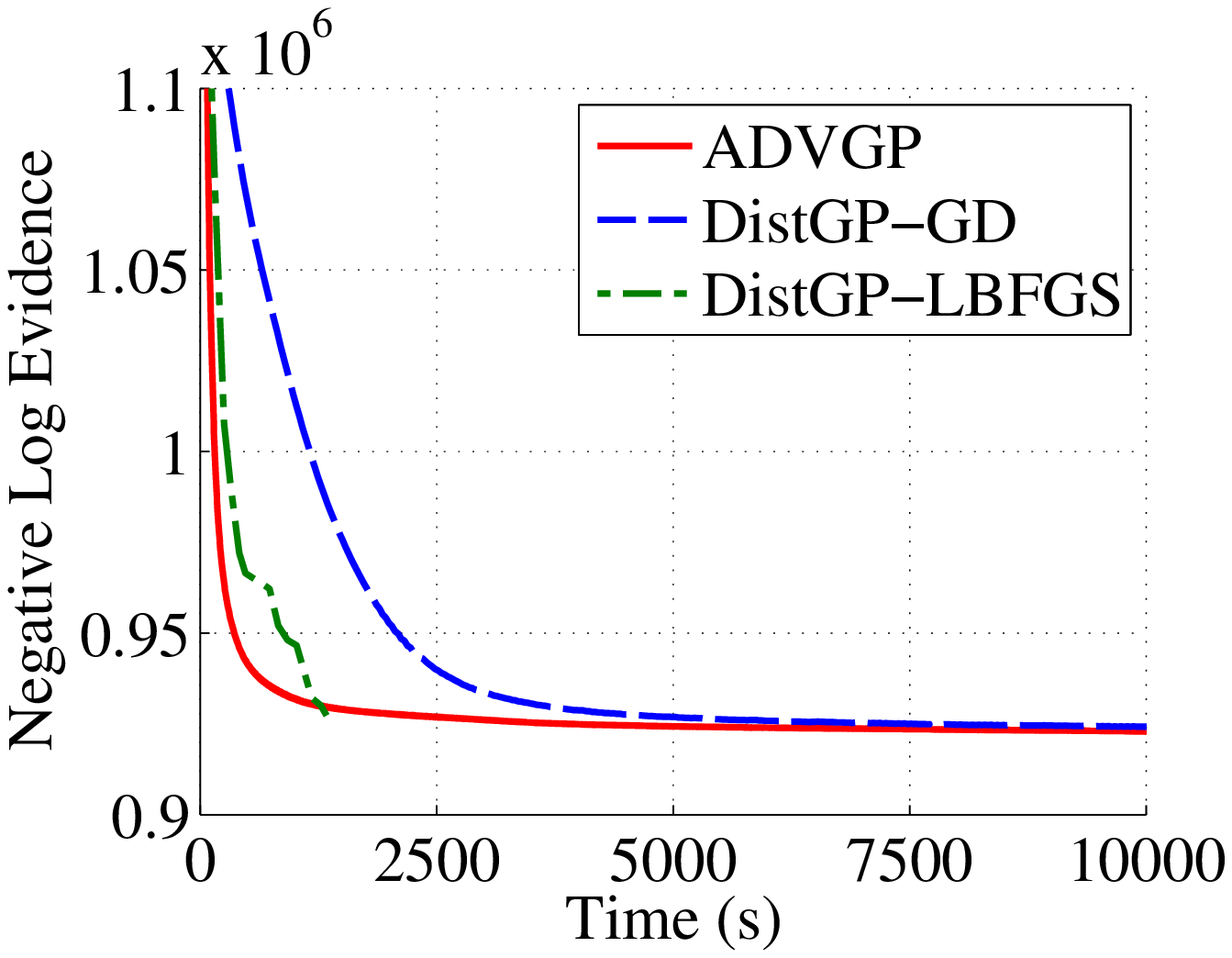}\\
(A) $m=100$ & (B)  $m=200$
\end{tabular}
\end{center}
\caption{Negative log evidences  for $700$K/$100$K US Flight data as a function of training time.} \label{fig:nle_700K}
\end{figure}

\begin{table}[ht!]
\begin{center}
\begin{tabular}{c || c | c}
Method  & $m = 100$ & $m = 200$ \\
\hline
ADVGP         & $\bf{925236}$ & $\bf{922907}$ \\
DistGP-GD    &  $927414$ & $924208$ \\
DistGP-LBFGS &  $932179$ & $927331$
\end{tabular}
\end{center}
\caption{Negative log evidences for $700$K/$100$K US Flight data.} 
\end{table}

\begin{figure}[ht!]
\begin{center}
\begin{tabular}{cc}
\includegraphics[width=0.45\linewidth]{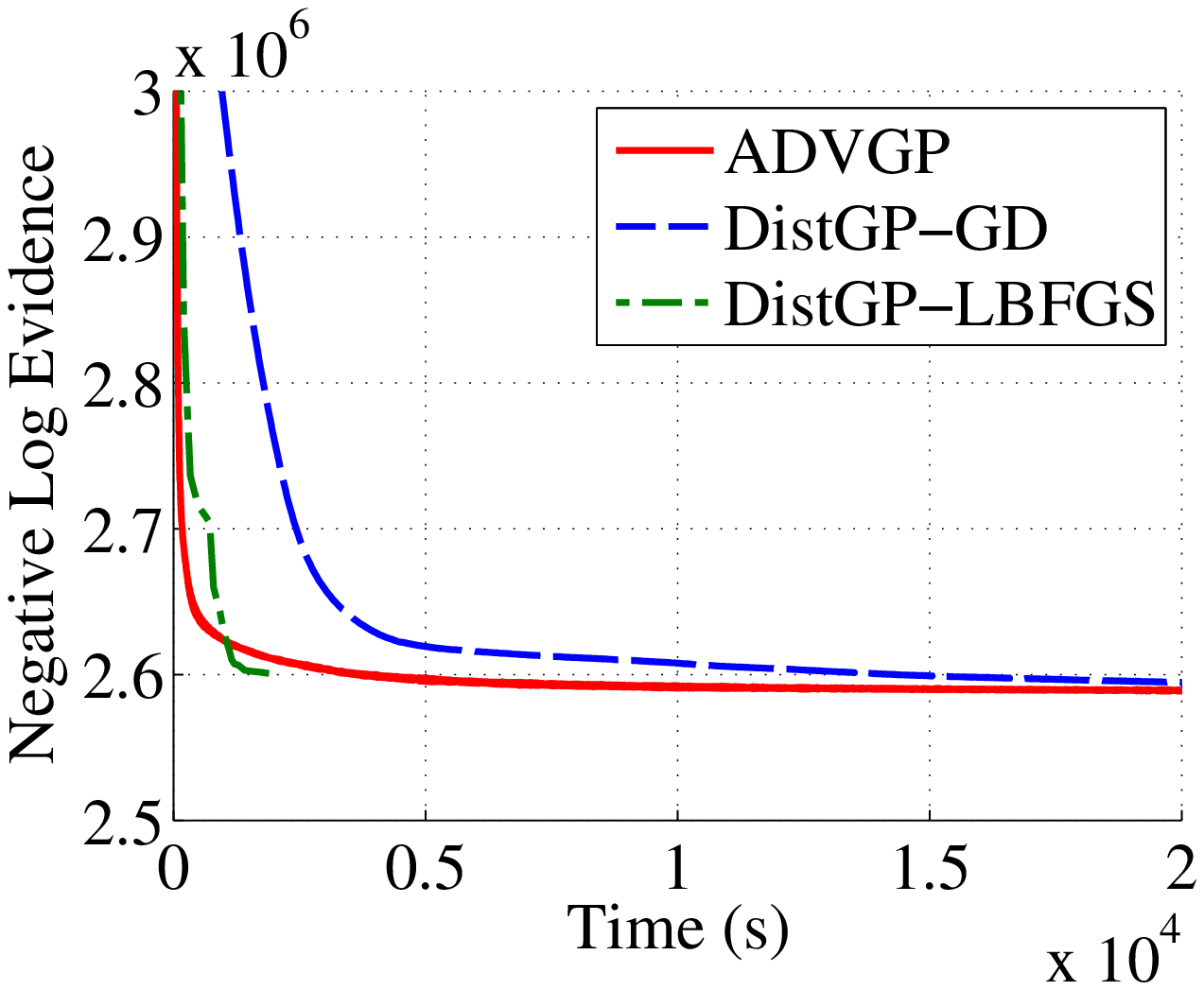}  &
\includegraphics[width=0.45\linewidth]{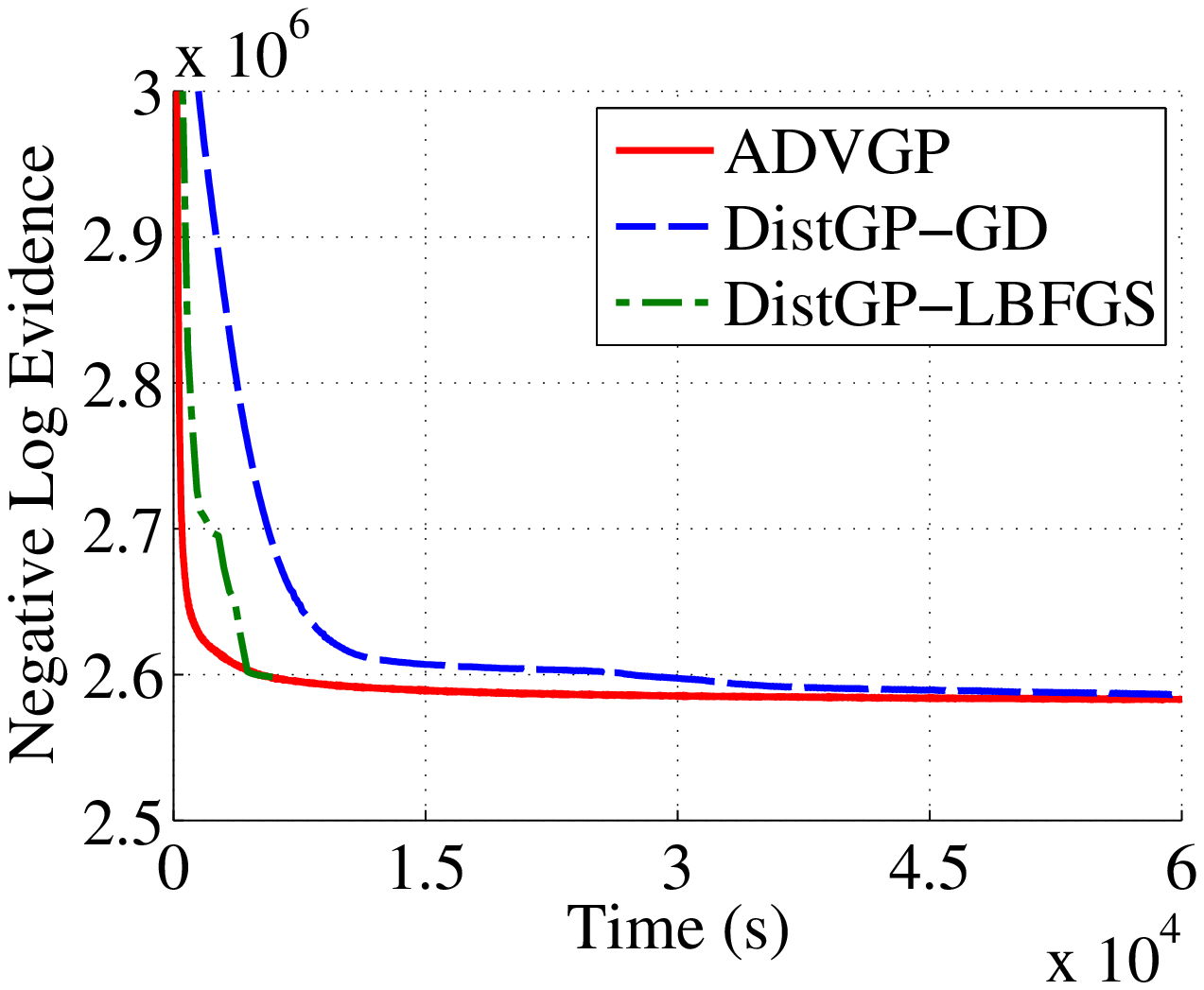}\\
(A) $m=100$ & (B)  $m=200$
\end{tabular}
\end{center}
\caption{Negative log evidences  for $2$M/$100$K US Flight data as a function of training time.} \label{fig:nle_2M}
\end{figure}

\begin{table}[ht!]
\begin{center}
\begin{tabular}{c || c | c}
Method  & $m = 100$ & $m = 200$ \\
\hline
ADVGP       & $\bf{2.58921\times10^6}$ & $\bf {2.58267\times10^6}$ \\
DistGP-GD    & $2.59471\times10^6$  & $2.58601\times10^6$ \\
DistGP-LBFGS & $2.59971\times10^6$ & $2.59817\times10^6$\\
\end{tabular}
\end{center}
\caption{Negative log evidences for $2$M/$100$K US Flight data.}
\end{table}

\newpage
\section{Mean Negative Log Predictive Likelihoods (MNLPs) on US Flight Data}
\begin{figure}[ht!]
\begin{center}
\begin{tabular}{cc}
\includegraphics[width=0.45\linewidth]{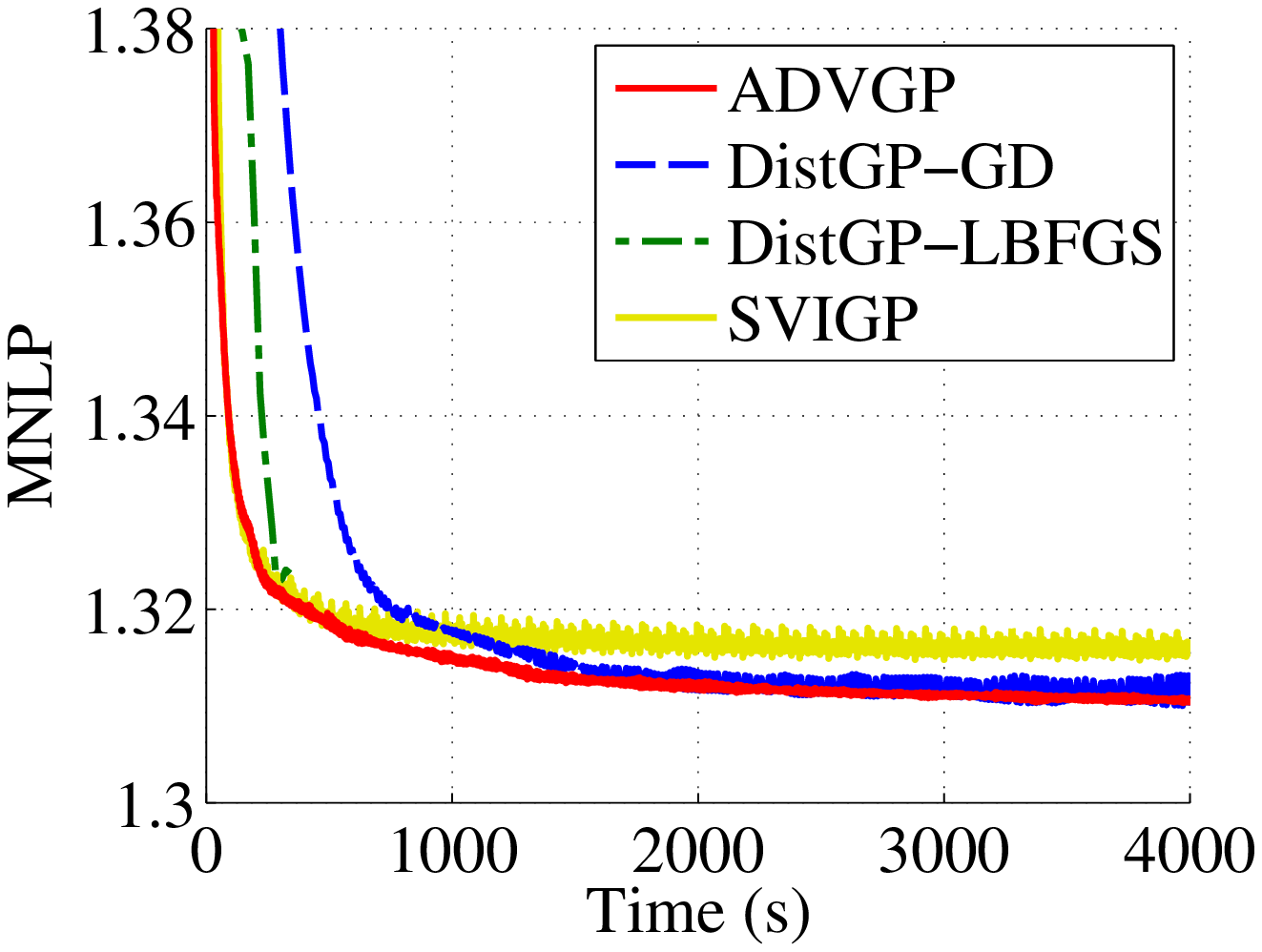}  &
\includegraphics[width=0.45\linewidth]{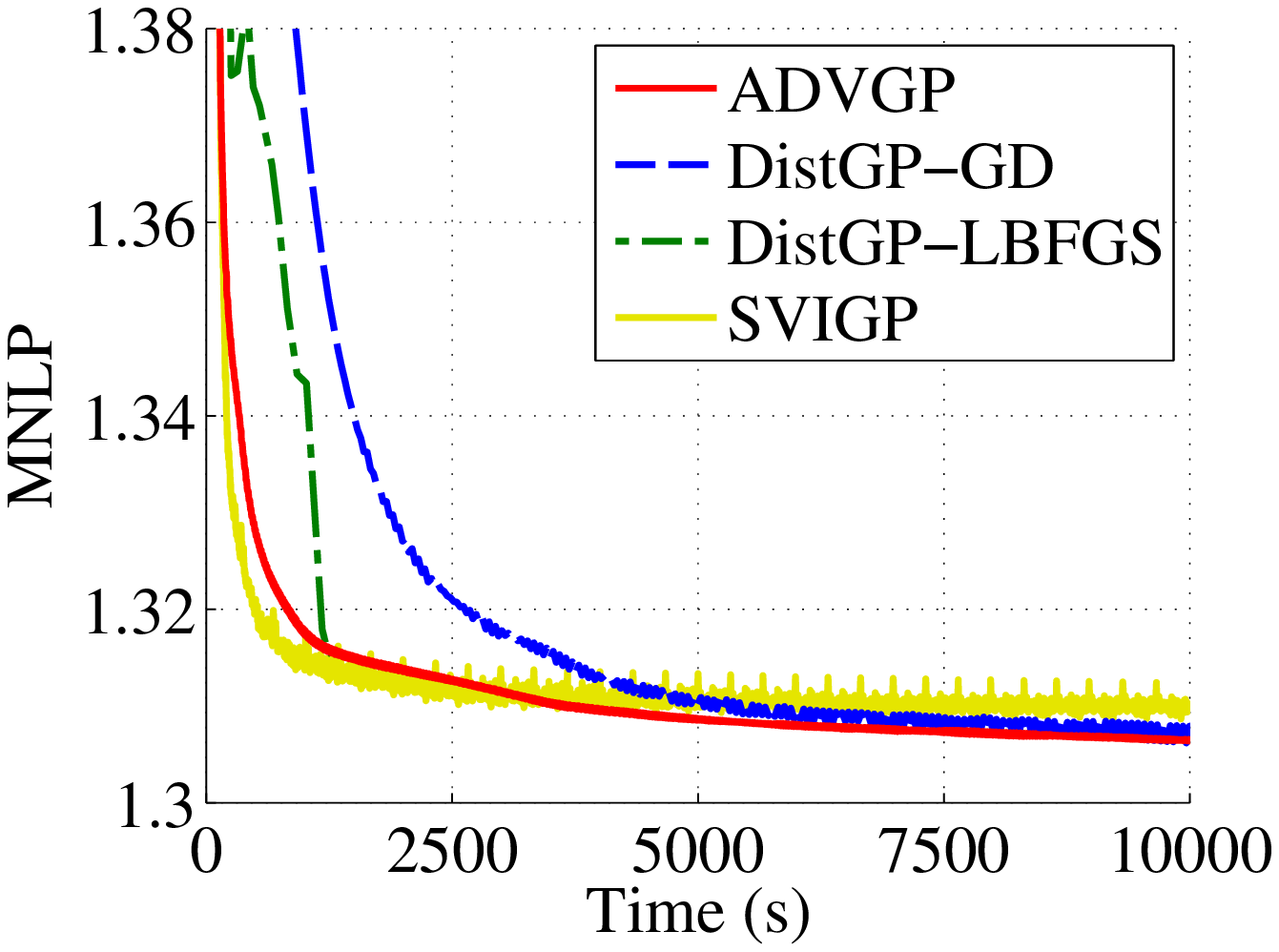}\\
(A) $m=100$ & (B)  $m=200$
\end{tabular}
\end{center}
\caption{Mean negative log predictive likelihoods for $700$K/$100$K US Flight data as a function of training time.} \label{fig:nle_2M}
\end{figure}

\begin{table}[ht!]
\begin{center}
\begin{tabular}{c || c | c}
Method  & $m = 100$ & $m = 200$ \\
\hline
ADVGP       & $1.3106$ & $1.3066$ \\
DistGP-GD    & $\bf{1.3099}$  & $\bf {1.3062}$ \\
DistGP-LBFGS & $1.3237$ & 1.3136\\
SVIGP     &  1.3157 &  1.3096 \\
\end{tabular}
\end{center}
\caption{Mean negative log predictive likelihoods for $700$K/$100$K US Flight data.} 
\end{table}

\begin{figure}[ht!]
\begin{center}
\begin{tabular}{cc}
\includegraphics[width=0.45\linewidth]{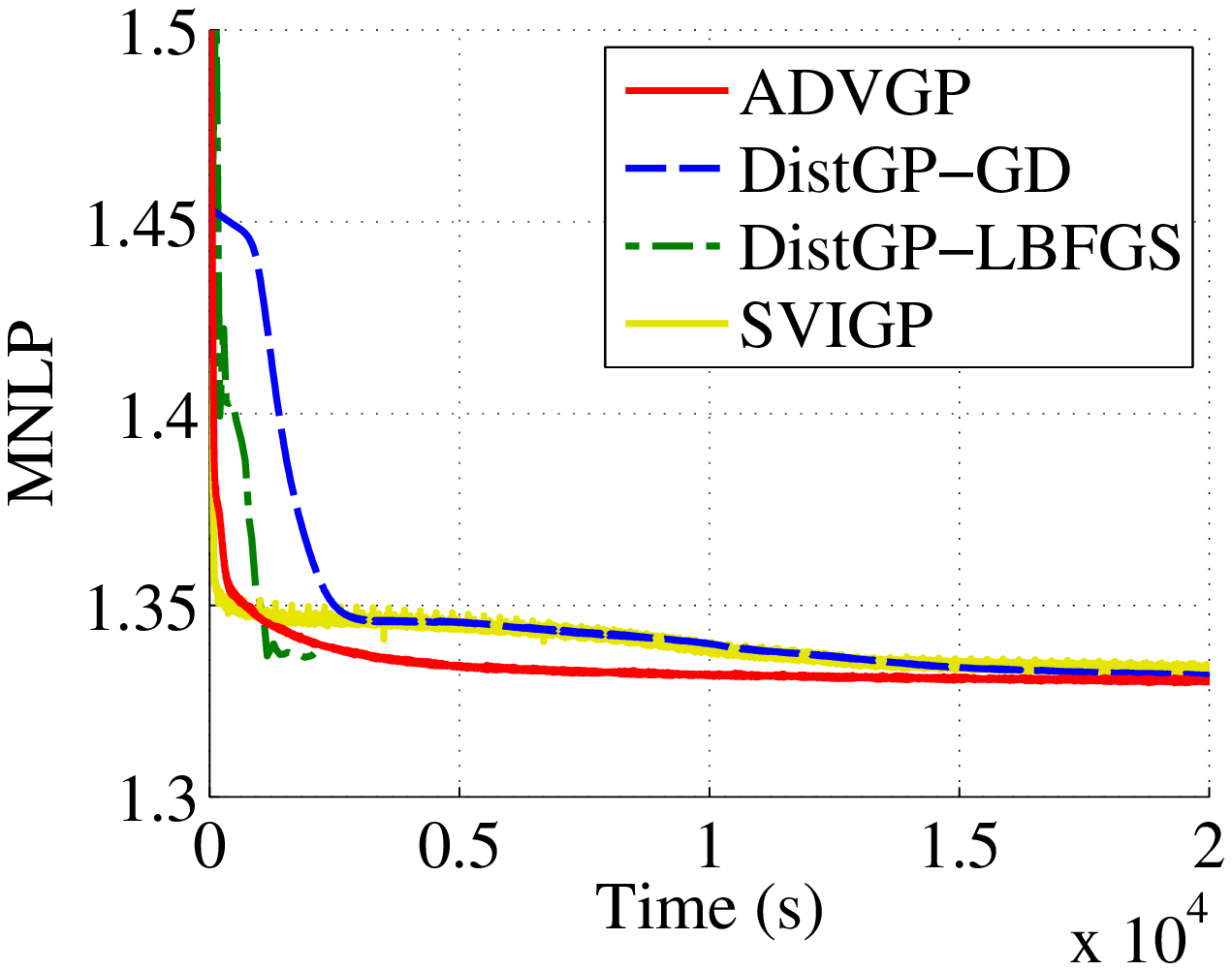}  &
\includegraphics[width=0.45\linewidth]{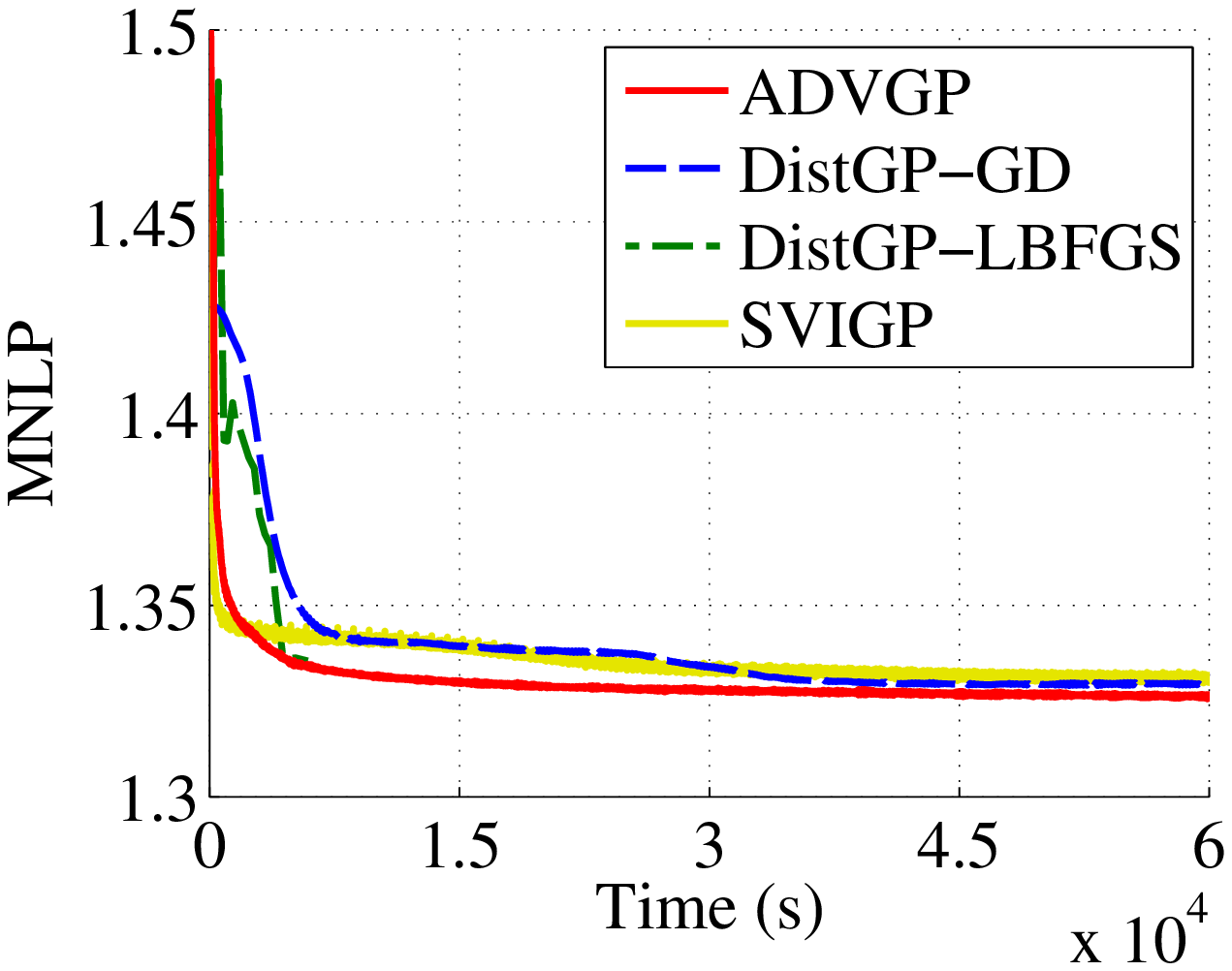}\\
(A) $m=100$ & (B)  $m=200$
\end{tabular}
\end{center}
\caption{Mean negative log predictive likelihoods for $2$M/$100$K US Flight data as a function of training time.} \label{fig:nle_2M}
\end{figure}

\begin{table}[ht!]
\begin{center}
\begin{tabular}{c || c | c}
Method  & $m = 100$ & $m = 200$ \\
\hline
ADVGP       & $\bf{1.3301}$ & $\bf {1.3258}$ \\
DistGP-GD    & $ 1.3317$  & $1.3297$ \\
DistGP-LBFGS & $1.3380$ & $1.3355$\\
SVIGP     &  1.3335 & 1.3306 \\
\end{tabular}
\end{center}
\caption{Mean negative log predictive likelihoods for $2$M/$100$K US Flight data.}
\end{table}


\end{document}